\documentclass[10pt,journal,compsoc]{IEEEtran}
\usepackage{cite}
\ifCLASSINFOpdf
   \usepackage[pdftex]{graphicx}
\else
\fi

\usepackage[utf8]{inputenc} 
\usepackage[T1]{fontenc}    
\usepackage{hyperref}       
\usepackage{url}            
\usepackage{booktabs}       
\usepackage{amsfonts}       
\usepackage{nicefrac}       
\usepackage{microtype}      
\usepackage{xcolor}         

\usepackage{graphicx}
\usepackage{amssymb}
\usepackage{pifont}

\usepackage[normalem]{ulem}

\newcommand{\cmark}{\ding{51}}%
\newcommand{\xmark}{\ding{55}}%

\usepackage{multirow}

\definecolor{red}{rgb}{.9,0.1,0.1}

\usepackage{wrapfig}
\usepackage{soul}

\begin{document}
%
\title{PASTA-GAN++: A Versatile Framework for High-Resolution Unpaired Virtual Try-on}
%
%
%
%

\author{Zhenyu~Xie, 
        Zaiyu~Huang,
        Fuwei~Zhao, 
        Haoye~Dong, 
        Michael~Kampffmeyer, 
        Xin~Dong, 
        Feida~Zhu, 
        Xiaodan~Liang,~\IEEEmembership{Senior Member,~IEEE}
\IEEEcompsocitemizethanks{\IEEEcompsocthanksitem X. Liang is the corresponding author. E-mail: xdliang328@gmail.com. Our code will be available at~\href{https://github.com/xiezhy6/PASTA-GAN-plusplus}{PASTA-GAN++}.\protect\\
\IEEEcompsocthanksitem Z. Xie, Z. Huang, F. Zhao and X. Liang are with Shenzhen Campus of Sun Yat-sen University, China. H. Dong is with Carnegie Mellon University, America. M. Kampffmeyer is with UiT The Arctic University of Norway, Norway. X. Dong and F. Zhu are with ByteDance. This work is done during Z. Xie's internship at ByteDance.
\IEEEcompsocthanksitem This work extends our previous work~\cite{xie2021pastagan} in four ways: (1) It devises a more dedicated patch representation for the various garments and (2) introduces a patch-guided parsing synthesis block into the dual-path Stylegan2 conditional generator, which enable the model to handle arbitrary garment categories and support fine-grained garment editing. (3) It conducts extensive experiments on a high-resolution benchmark to demonstrate its superiority over existing SOTAs, and (4) displays a variety of applications it supports to illustrate its versatility for controllable garment editing. 
}
}

\markboth{IEEE TRANSACTIONS ON PATTERN ANALYSIS AND MACHINE INTELLIGENCE, VOL.,No. XX, XX 2022}
{Shell \MakeLowercase{\textit{et al.}}: Bare Demo of IEEEtran.cls for Computer Society Journals}
%



\IEEEtitleabstractindextext{%
\begin{abstract}
Image-based virtual try-on is one of the most promising applications of human-centric image generation due to its tremendous real-world potential. 
In this work, we take a step forwards to explore versatile virtual try-on solutions, which we argue should possess three main properties, namely, they should support unsupervised training, arbitrary garment categories, and controllable garment editing.
To this end, we propose a characteristic-preserving end-to-end network, the PAtch-routed SpaTially-Adaptive GAN++ (PASTA-GAN++), to achieve a \emph{versatile} system for \emph{high-resolution unpaired} virtual try-on.
Specifically, our PASTA-GAN++ consists of an innovative patch-routed disentanglement module to decouple the intact garment into normalized patches, which is capable of retaining garment style information while eliminating the garment spatial information, thus alleviating the overfitting issue during unsupervised training.
Furthermore, PASTA-GAN++ introduces a patch-based garment representation and a patch-guided parsing synthesis block, allowing it to handle arbitrary garment categories and support local garment editing.
Finally, to obtain try-on results with realistic texture details, PASTA-GAN++ incorporates a novel spatially-adaptive residual module to inject the coarse warped garment feature into the generator. 
Extensive experiments on our newly collected UnPaired virtual Try-on (UPT) dataset demonstrate the superiority of PASTA-GAN++ over existing SOTAs and its ability for controllable garment editing.
\end{abstract}

}

\maketitle

\IEEEdisplaynontitleabstractindextext

%
\IEEEpeerreviewmaketitle

\IEEEraisesectionheading{\section{Introduction}\label{sec:introduction}}

%
%
%
%

\begin{table*}[t]
    \caption{Comparison between PASTA-GAN++ and existing methods in terms of the training data, image resolution, whether supporting arbitrary garment try-on, and whether supporting garment editing (i.e., dressing order control (DOC), style/texture transfer (STT), and local shape editing (LSE)).}
    \def\arraystretch{0.9}
    \small
    \tabcolsep 13pt
    \centering
    \begin{tabular}{l|c|c|c|ccc}
        \toprule
        \multirow{2}*{Method} & \multirow{2}*{Training Data}  & \multirow{2}*{Image Resolution} & \multirow{2}*{Arbitrary Category} & \multicolumn{3}{c}{Garment Editing} \\
        \cmidrule{5-7}
        & & & & DOC & STT & LSE \\
        \midrule
        VITON~\cite{xintong2018viton}   & Paired Data    & 256$\times$192    & \xmark   & \xmark  & \xmark    & \xmark  \\
        CP-VTON~\cite{bochao2018cpvton}   & Paired Data    & 256$\times$192    & \xmark   & \xmark  & \xmark    & \xmark  \\
        ACGPN~\cite{han2020acgpn}         & Paired Data    & 256$\times$192    & \xmark   & \xmark  & \xmark    & \xmark  \\
        VITON-HD~\cite{choi2021vitonhd}         & Paired Data    & 1024$\times$768    & \xmark   & \xmark  & \xmark    & \xmark  \\
        PF-AFN~\cite{ge2021pfafn}         & Paired Data    & 256$\times$192    & \xmark   & \xmark  & \xmark    & \xmark  \\
        RT-VTON~\cite{yang2022rtvton}         & Paired Data    & 256$\times$192    & \xmark   & \xmark  & \xmark    & \xmark  \\
        \midrule
        LWG~\cite{lwb2019}                & Paired Data    & 256$\times$256    & \xmark   & \xmark  & \xmark    & \cmark  \\
        ADGAN~\cite{men2020adgan}         & Paired Data    & 256$\times$176    & \cmark   & \xmark  & \cmark    & \xmark  \\
        Dior~\cite{Cui2021dior}           & Paired Data    & 256$\times$176    & \cmark   & \cmark  & \cmark    & \xmark  \\
        StylePoseGAN~\cite{sarkar2021posestylegan}        & Paired Data    & 512$\times$384    & \xmark   & \xmark  & \xmark    & \cmark  \\
        PWS~\cite{albahar2021pose}        & Paired Data    & 512$\times$384    & \cmark   & \xmark  & \xmark    & \cmark  \\
        wFlow~\cite{dong2022wflow}        & Paired Data    & 512$\times$512    & \cmark   & \xmark  & \xmark    & \xmark  \\
        \midrule
        O-VITON~\cite{neuberger2020oviton} & Unpaired Data   & 512$\times$256    & \cmark   & \xmark  & \xmark    & \xmark  \\
        TryOnGAN~\cite{ira2021vogue}      & Unpaired Data   & 512$\times$512    & \xmark   & \xmark  & \xmark    & \xmark  \\
        \midrule
        PASTA-GAN~\cite{xie2021pastagan}  & Unpaired Data   & 256$\times$192    & \xmark   & \xmark  & \cmark    & \cmark  \\
        PASTA-GAN++                       & Unpaired Data   & 512$\times$320    & \cmark   & \cmark  & \cmark    & \cmark  \\
        \bottomrule
    \end{tabular}
    \label{tab:try-on methods comparison}
\end{table*}

\begin{figure*}[t]
  \centering
  \includegraphics[width=1.0\hsize]{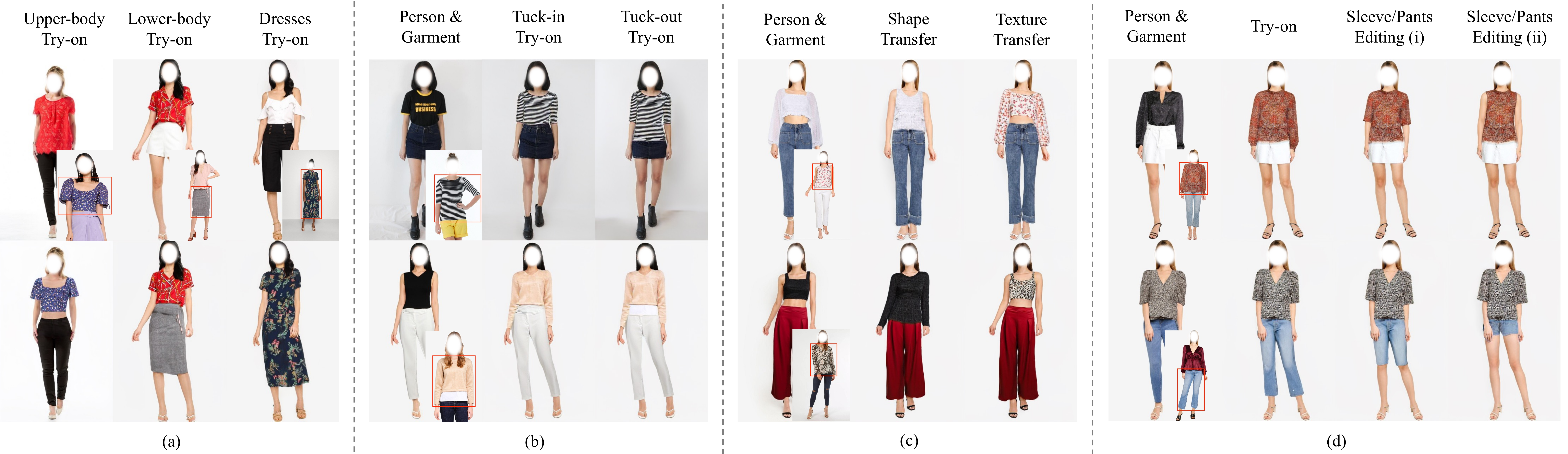}
  \caption{Our PASTA-GAN++ is a versatile framework for unpaired virtual try-on, which supports (a) arbitrary garment try-on (e.g. upper-body, lower-body, and dresses try-on) and several types of garment editing including (b) dressing order control, (c) garment shape/texture transfer, (d) local shape editing. Please zoom in for more details.}
  \label{fig:teaser}
\end{figure*}

\IEEEPARstart{I}{mage-based} virtual try-on, the process of computationally transferring a garment onto a particular person in a query image, is one of the most promising applications of human-centric image generation with the potential to revolutionize shopping experiences and reduce purchase returns. 
The computer vision community has recently witnessed the increasing development of image-based virtual try-on, and numerous impressive works~\cite{xintong2018viton,bochao2018cpvton,han2019clothflow,han2020acgpn,ge2021pfafn,yang2022rtvton,he2022fs_vton,lee2022hrviton,xie2021pastagan,albahar2021pose,dong2022wflow} have been proposed to synthesize photo-realistic try-on results on the publicly available benchmarks~\cite{xintong2018viton,dong2019mgvton,choi2021vitonhd,liuLQWTcvpr16DeepFashion}.
However, to fully exploit its potential, versatile virtual try-on solutions are required, which possess the following three properties: First, the training procedure of the virtual try-on network should make full use of the easily accessible \emph{unpaired} fashion model images from e-commerce websites, which means that the network should be easily trainable in an unsupervised manner. Second, a versatile virtual try-on solution should be capable of handling arbitrary garment categories (e.g., long/short sleeve shirt, vest, sling, pants, shorts, skirts, dresses, etc.) within a single pre-trained model. Third, it needs to support auxiliary applications (e.g., dressing order controlling, shape/texture editing, local shape editing, etc.) to fulfill various requirements of dressing styles or garment editing in real-world scenarios. 
Unfortunately, as shown in Table~\ref{tab:try-on methods comparison}, to date, few methods can achieve all of the above three requirements simultaneously. 
Most existing methods~\cite{xintong2018viton,bochao2018cpvton,han2020acgpn,choi2021vitonhd,ge2021pfafn,yang2022rtvton,lwb2019,Cui2021dior,men2020adgan,albahar2021pose,dong2022wflow} rely on \emph{paired} training data from curated academic datasets~\cite{xintong2018viton,dong2019mgvton,choi2021vitonhd,liuLQWTcvpr16DeepFashion}, in which each data pair is composed of a person image and its corresponding in-shop garment or of images of the same person captured in different body poses, resulting in laborious data-collection and post-processing processes. Besides, most existing methods~\cite{xintong2018viton,bochao2018cpvton,han2020acgpn,choi2021vitonhd,ge2021pfafn,yang2022rtvton} only focus on upper body virtual try-on and fail to support arbitrary categories within a single pre-trained model. 
Furthermore, most existing methods~\cite{lwb2019,men2020adgan,albahar2021pose,neuberger2020oviton,ira2021vogue} neglect the dressing styles and lack the ability to edit the garment texture/shape, which largely limits their application potential.

While \emph{unpaired} solutions have recently started to emerge, performing virtual try-on in an unsupervised setting is extremely challenging and tends to affect the visual quality of the try-on results.
Specifically, without access to the paired data, these models are usually trained by reconstructing the same person image, which is prone to over-fitting, and they thus underperform when handling garment transfer during testing. The performance discrepancy is mainly reflected in the garment synthesis results, in particular the shape and texture, which we argue is caused by the entanglement of the garment style (i.e., color, category) and spatial (i.e.,the location, orientation, and relative size of the garment in the model image) representations in the synthesis network during the reconstruction process. 

While traditional paired try-on approaches, such as the warping-based methods~\cite{bochao2018cpvton,han2019clothflow,han2020acgpn,ge2021pfafn,Cui2021dior,albahar2021pose,dong2022wflow} avoid the problem and preserve the garment characteristics by utilizing a supervised warping network to deform the garment into target shape, this is not possible in the unpaired setting due to the lack of the warped ground truth. Similarly, warping-free methods~\cite{men2020adgan,sarkar2020nhrr,sarkar2021humangan,sarkar2021posestylegan}, which choose to circumvent this problem by using person images in various poses as training data and taking pose transfer as pretext task to disentangle the garment feature from the intrinsic body pose, also require a laborious data-collection process for paired data, largely limiting the scalability of network training. 
The few works~\cite{neuberger2020oviton,ira2021vogue} that attempt to achieve unpaired virtual try-on train an unsupervised try-on network and then exploit extensive online optimization procedures to obtain fine-grained details of the original garments, harming the inference efficiency. Furthermore, none of the existing unpaired try-on methods consider the problem of coupled style and spatial garment information directly, which is crucial to obtain accurate garment transfer results in the unpaired and unsupervised virtual try-on scenario.

On the other hand, the choice of garment representation is crucial in devising a controllable virtual try-on network. A flexible garment representation can empower the network to accomplish fine-grained garment editing. Nevertheless, most existing methods~\cite{xintong2018viton,bochao2018cpvton,han2020acgpn,ge2021pfafn,yang2022rtvton} take intact garments as inputs and only focus on garment transfer, failing to conduct controllable garment editing. Some methods~\cite{men2020adgan,Cui2021dior} introduce human parsing into the try-on network and support fundamental editing like shape/texture transfer between two garments. However, since the human parsing is defined on an object level, it is unable to distinguish different regions (e.g., upper/lower sleeve region, upper/lower pants region, torso region, etc.) within the garment, which makes it challenging to conduct more fine-grained garment editing like changing the shape of a particular local region. Some advanced methods~\cite{lwb2019,sarkar2021posestylegan,albahar2021pose} exploit the pre-defined body segmentation from the 3D model~\cite{matthew2015smpl,guler2018densepose} to divide the garment into body-related patches, which enables the network to conduct editing at a patch-level. However, since the 3D model~\cite{matthew2015smpl,guler2018densepose} can only represent the human body without clothing, garment patches derived from the 3D model can not guarantee the completeness of the original garment, especially for loose garments. 
Therefore, in order to fulfill controllable editing during the virtual try-on process, a feasible garment representation is required, which not only separates the garment into fine-grained patches but also maintains the garment's completeness.

In this paper, to tackle the essential challenges mentioned above and resolve the deficiency of prior approaches, 
we propose a novel PAtch-routed SpaTially-Adaptive GAN++ (PASTA-GAN++), a versatile solution to the \emph{high-resolution unpaired} virtual try-on task. Our PASTA-GAN++ can precisely synthesize garment shape and texture by introducing a patch-routed disentanglement module that decouples the garment style and spatial features, a patch-guided parsing synthesis block to generate correct garment shapes complying with specific body poses, as well as a novel spatially-adaptive residual module to mitigate the problem of feature misalignment. Besides, due to the well-designed garment 
patches, our PASTA-GAN++ can handle arbitrary garment categories (e.g., shirt, vest, sling, pants, skirts, dresses, etc.) through a single pre-trained model and further supports garment editing (e.g. dressing order controlling, shape/texture transfer, local shape editing, etc.) during the try-on procedure (see Fig.~\ref{fig:teaser}). 

The innovation of our PASTA-GAN++ includes four aspects: 
First, by separating the garments into normalized patches with the inherent spatial information largely reduced, the patch-routed disentanglement module encourages the style encoder to learn spatial-agnostic garment features. These features enable the synthesis network to generate images with accurate garment style regardless of varying spatial garment information.
Second, the well-designed garment patches provide a fine-grained and unified representation for garments from various categories, which is leveraged by the patch-guided parsing synthesis block to generate correct target shape for arbitrary garment category within a single pre-trained model, and enables the model to conduct fine-grained garment editing.
Third, given the target human pose, the normalized patches can be easily reconstructed to the warped garment complying with the target shape, without requiring a warping network or a 3D human model. 
Finally, the spatially-adaptive residual module extracts the warped garment feature and adaptively inpaints the region that is misaligned with the target garment shape. Thereafter, the inpainted warped garment features are embedded into the intermediate layer of the synthesis network, guiding the network to generate try-on results with realistic garment texture. 

To explore the proposed high-resolution unpaired virtual try-on algorithm, we collect a large number of high-quality images 
and combine them with a 
subset of the existing virtual try-on benchmarks~\cite{liuLQWTcvpr16DeepFashion,dong2019mgvton} to construct a scalable UnPaired virtual Try-on (UPT) dataset, which contains more than 100k high-resolution front-view fashion model images wearing a large variety of garments, e.g., long/short sleeve shirt, sling, vest, sling, pants, shorts, skirts, dresses, etc. Extensive experiment results on the UPT dataset demonstrate that our unsupervised PASTA-GAN++ outperforms the previous virtual try-on approaches and can obtain impressive results for controllable garment editing. 

\section{Related Work}

\subsection{Human-centric Image Synthesis}
Recently, the promising performance of StyleGAN-based models ~\cite{karras2019stylegan,karras2020stylegan2,karras2020stylegan2ada,karras2021stylegan3} to generate high-fidelity images has been demonstrated. They are able to generate high-resolution photo-realistic images for individual classes (e.g., human face, animal, car, etc.) and can be regarded as the state-of-the-art framework for unconditional image synthesis. 
Besides, due to the inherent multigrain semantic information in the StyleGAN latent space, StyleGAN-based generator can further be leveraged for controllable image editing through modifying the latent code in its learnable latent space.
Therefore, most of the existing human-centric image synthesis methods~\cite{anna2022insetgan,fu2022styleganhuman,sarkar2021posestylegan,albahar2021pose} inherit the StyleGAN-based architecture for high-fidelity human synthesis. InsetGAN~\cite{anna2022insetgan} proposes a novel generation mechanism to combine multiple pretrained GANs for full-body human synthesis, in which one GAN aims at global body synthesis while the other GANs focus on various local part synthesis (e.g., face, hands, etc.). With such a global-local composition mechanism, InsetGAN is capable of generating full-body human images with dedicated details in the local body parts (e.g., face, hands, etc.). StyleGAN-Human~\cite{fu2022styleganhuman} re-thinks the StyleGAN-based human image synthesis from a "data engineering" perspective and explores three crucial factors for high-quality human image synthesis, namely, data size, data distribution, and data alignment. 
Although these methods~\cite{anna2022insetgan,fu2022styleganhuman} are well-designed for photo-realistic human synthesis, they can only generate random images and fail to accomplish image synthesis according to specific conditions, which is a crucial requirement for the virtual try-on scenario.

In contrast to the above unconditional synthesis models,
some advanced methods~\cite{sarkar2021posestylegan,albahar2021pose} are developed for conditional human synthesis, which are trained in a supervised manner by using paired data (i.e., images of the same person under different poses.). Specifically, StylePoseGAN~\cite{sarkar2021posestylegan} conditions the StyleGAN generator on the human densepose~\cite{guler2018densepose} by replacing the constant input of the generator with a 2D pose feature map extracted from the person densepose. Besides, the 1D appearance feature derived from the texture UV map is regarded as the latent code to modulate the weights of 
convolution layers in the generator. Pose-with-style~\cite{albahar2021pose} also devises a densepose-guided StyleGAN generator. 
However, instead of directly using the appearance feature for the UV map, Pose-with-style applies an appearance flow, which is estimated using a novel coordinate completion module, to warp the source appearance feature to the target pose and uses the 2D warped feature to modulate the generator.
Although these modified StyleGANs can accomplish conditional image synthesis, their training procedures rely on paired data, thus requiring laborious dataset collection and violating the task setting of unpaired virtual try-on.

\begin{figure*}[t]
  \centering
  \includegraphics[width=1.0\hsize]{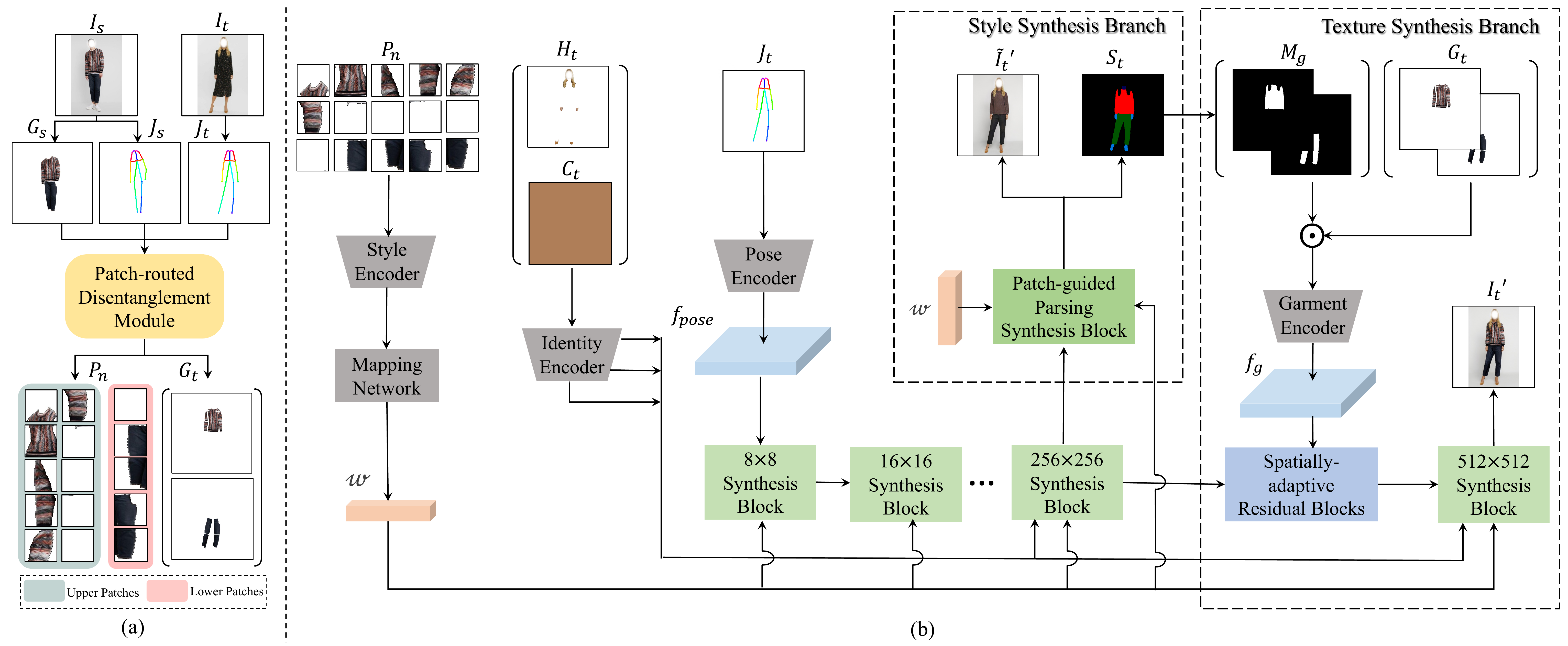}
 \vspace{-4mm}
  \caption{Overview of the inference process. (a) Given the source and target images of person $(I_s, I_t)$, we can first extract the source garment $G_s$, the source pose $J_s$, and the target pose $J_t$. The three are then sent to the patch-routed disentanglement module to yield the normalized garment patches $P_n$ (including upper and lower patches) and the warped garments $G_t$ (including upper and lower warped garment). (b) The dual-path conditional StyleGAN2 first collaboratively exploits the disentangled style code $w$, projected from $P_n$, the pose feature $f_{pose}$, encoded from target pose $J_t$, the multi-scale identity feature, encoded from the preserved region and median skin color map $(H_t, C_t)$ to synthesize the coarse try-on result $\tilde{I_t}'$ in the style synthesis branch along with the target human parsing $S_t$. It then leverages the warped garment feature $f_g$ in the texture synthesis branch to generate the final try-on result $I_t'$.}
  \vspace{-4mm}
  \label{fig:framework}
\end{figure*}

\subsection{Image-based Virtual Try-on}
Given the reference person and the target garment, image-based virtual try-on aims to synthesize a photo-realistic image of a person wearing the specified garment. Most of the existing works~\cite{xintong2018viton,bochao2018cpvton,han2019clothflow,han2020acgpn,xie2021wasvton,ge2021pfafn,yang2022rtvton,men2020adgan,lwb2019,sarkar2021posestylegan,albahar2021pose,dong2022wflow} explore virtual try-on  under the paired setting and employ paired data for network training, which limits their generalization to easily accessible unpaired fashion images from e-commerce websites. However, there exist a few works~\cite{neuberger2020oviton,ira2021vogue} designed for the unpaired setting, which aim to avoid the laborious paired training data collection process. In the following, we will summarize the typical paired and unpaired virtual try-on approaches, respectively.

\textbf{Paired virtual try-on.}
Most of the existing paired virtual try-on methods~\cite{xintong2018viton,bochao2018cpvton,yun2019vtnfp,han2019clothflow,Matiur2020cpvton+,han2020acgpn,thibaut2020wuvton,chongjian2021dcton,ge2021pfafn,xie2021wasvton,lin2022viton,yang2022rtvton,he2022fs_vton,lee2022hrviton,bai2022sdafn} follow the garment-to-person paradigm, which aims to transfer an in-shop garment onto a reference person. The paired training data for this paradigm is composed of a person image and its corresponding in-shop garment image. Among these methods, VITON~\cite{xintong2018viton} for the first time integrates a U-Net~\cite{ronneberger2015unet} based generation network with a TPS~\cite{bookstein1989TPS} based deformation approach to synthesize the try-on result. CP-VTON~\cite{bochao2018cpvton} proposes a two-stage framework to first deform the in-shop garment to the target shape and then synthesize the try-on result. Most of the following works aim to improve particular modules in this two-stage framework.
To improve the quality of the garment deformation, ~\cite{Matiur2020cpvton+,han2020acgpn,chongjian2021dcton,Zhao202m3dvton,yang2022rtvton} introduce particular smoothness constraints for the warping module to alleviate the excessive distortion in TPS warping, while\cite{han2019clothflow,xie2021wasvton,ge2021pfafn,he2022fs_vton,lee2022hrviton,bai2022sdafn} turn to flow-based warping schemes which models the per-pixel deformation.
To provide the final try-on module with explicit body shape guidance,
~\cite{yun2019vtnfp,han2020acgpn} adopt conditional human parsing to guide the generation of various body parts.
\cite{choi2021vitonhd,lee2022hrviton} instead focus on high-resolution virtual try-on and propose an ALIAS normalization mechanism to resolve the garment misalignment, while ~\cite{thibaut2020wuvton,ge2021pfafn,lin2022viton} improve the learning process by employing knowledge distillation, resulting in a parser-free virtual try-on framework.
However, all of these methods require paired training data and are incapable of providing controllable garment editing during the try-on procedure.

On the other hand, there exists another kind of paired solutions~\cite{lwb2019,men2020adgan,sarkar2020nhrr,sarkar2021humangan,sarkar2021posestylegan,Cui2021dior,albahar2021pose,dong2022wflow}, namely following the person-to-person paradigm, which directly exchanges garments between two input person images. The paired training data for this paradigm consists of multiple images of the same person in different body poses. These methods generally exploit pose transfer as the pretext task to decouple the appearance and pose feature for human synthesis~\cite{men2020adgan,sarkar2020nhrr,sarkar2021humangan,sarkar2021posestylegan} or learn the appearance deformation function between the source and target poses~\cite{Cui2021dior,albahar2021pose,dong2022wflow}. During testing, virtual try-on can be accomplished by combining the synthesized garment that complies with the target pose and the remaining body parts of the target person. Similarly, the training procedures of these methods still rely on the hard-to-collect paired data.

\textbf{Unpaired virtual try-on.} Unpaired methods
~\cite{neuberger2020oviton,ira2021vogue} are more flexible and can be directly trained with unpaired person images.
However, OVITON~\cite{neuberger2020oviton} requires online appearance optimization for each garment region during testing to maintain texture detail of the original garment and VOGUE~\cite{ira2021vogue} needs to separately optimize the latent codes for each person image and then interpolate coefficients for the final try-on result during testing.
Therefore, existing unpaired methods require extensive online optimization, extremely harming their scalability in real scenarios. 

\begin{figure*}[t]
\centering
\includegraphics[width=\linewidth]{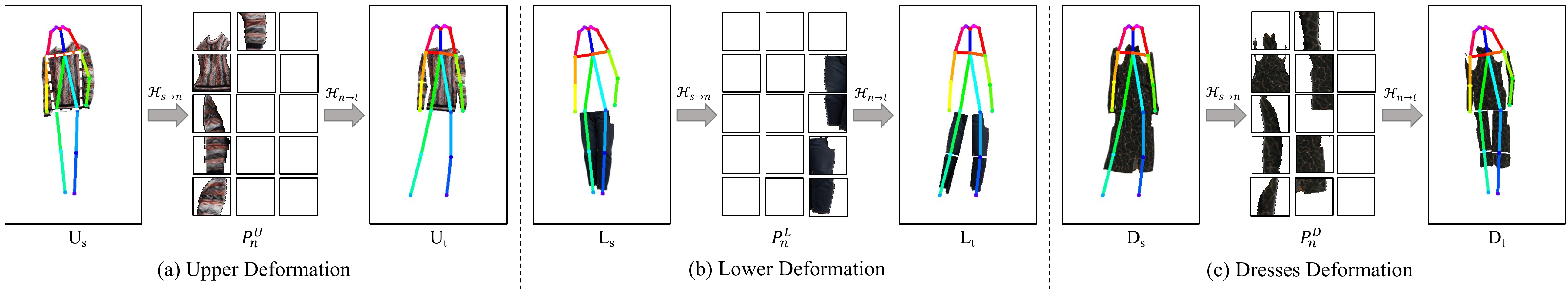}
\caption{The patch-routed deformation process for upper-body garments, lower-body garments, and dresses. Please zoom in for more details.}
\vspace{-3mm}
\label{fig:patch-routed}
\end{figure*}

\section{PASTA-GAN++}
Given a source image $I_s$ of a person wearing a garment $G_s$, and a target person image $I_t$, the unpaired virtual try-on task aims to synthesize the try-on result $I_t'$ retaining the identity of $I_t$ but wearing the source garment $G_s$. 
To achieve this, our PASTA-GAN++ first utilizes the patch-routed disentanglement module (Sec.~\ref{tab:sec3-1}) to transform the garment $G_s$ into normalized patches $P_n$ that are mostly agnostic to the spatial features of the garment, and further deforms $P_n$ to obtain the warped garment $G_{t}$ complying with the target person pose. 
Then, a dual path conditional StyleGAN2 (Sec.~\ref{tab:sec3-2}) is designed to synthesize the photo-realistic try-on result with correct garment shape and texture, where we introduce a patch-guided parsing synthesis block to generate a human parsing $S_t$ describing the precise shape of various garments and the novel spatially-adaptive residual blocks (Sec.~\ref{tab:sec:3-3}) to inject the warped garment features into the generator network for more realistic texture synthesis. The loss functions and training details will be described in Sec.~\ref{tab:sec3-4}. 
Fig.~\ref{fig:framework} illustrates the overview of the inference process for PASTA-GAN++.

\subsection{Patch-routed Disentanglement Module}~\label{tab:sec3-1}
Since the paired data for supervised training is unavailable for the unpaired virtual try-on task, the synthesis network has to be trained in an unsupervised manner via image reconstruction,
and thus takes a person image as input and separately extracts the feature of the intact garment and the feature of the person representation to reconstruct the original person image.
While such a training strategy retains the intact garment information, which is helpful for the garment reconstruction, the features of the intact garment entangle the garment style with the spatial information in the original image. This is detrimental to the garment transfer during testing.
Note that the garment style here refers to the garment color and categories, i.e., long sleeve, short sleeve, etc., while the garment spatial information implies the location, the orientation, and the relative size of the garment patch in the person image, in which the first two parts are influenced by the human pose while the third part is determined by the relative camera distance to the person.

To address this issue, we propose a novel patch-based garment representation for the unpaired virtual try-on model, in which the intact garment is explicitly divided into normalized patches to remove the inherent spatial information of the garment. Taking the sleeve patch as an example, by using division and normalization, various sleeve regions from different person images can be deformed to normalized patches with the same orientation and scale. Without the guidance of the spatial information, the network is forced to learn the garment style feature to reconstruct the garment in the synthesis image. 
Furthermore, to provide a unified representation for arbitrary garments, the garment patches are further divided into two kinds of patches, namely, a set of upper patches and a set of lower patches, where the upper patches are used to represent upper-body garments (e.g., shirt, coat, etc.) and dresses, while the lower-body garments (e.g., pants, skirts, etc.) are represented by the lower patches. The upper and lower patches consist of ten patches (i.e., four patches around the left/right upper/bottom arm, four patches around the left/right upper/bottom leg, a patch around the torso, and a patch around the neck) and five patches (i.e., four patches around the left/right upper/bottom leg, and a patch around the torso), respectively.

Fig.~\ref{fig:patch-routed} illustrates the process of obtaining normalized garment patches for upper-body garments, lower-body garments, and dresses, each of which includes two main steps: (1) pose-guided garment segmentation, and (2) perspective transformation-based patch normalization. 
Specifically, in the first step, the source garment $G_s$ (i.e., $U_s$, $L_s$, and $D_s$) and human pose (joints) $J_s$ are firstly obtained by applying~\cite{Gong2019Graphonomy} and~\cite{openpose} to the source person $I_s$, respectively. Given the body joints, we can segment the source garment $G_s$ into several patches $P_{n}$, which can be quadrilaterals with arbitrary shapes (e.g., rectangle, square, trapezoid, etc.), and will later be normalized.
Taking the torso region of an upper-body garment as an example, with the coordinates of the left/right shoulder joints and the left/right hips joints in $J_s$, a quadrilateral crop covering the torso region of $G_s$ can be easily performed to produce an unnormalized garment patch $P_s^i$ (i.e., the white dotted box in Fig.~\ref{fig:patch-routed}(a)).
In the second step, all patches are normalized to remove their spatial information by perspective transformations. For this, we first define the same amount of template patches $P_{n}$ with fixed $128\times128$ resolution as transformation targets for all unnormalized source patches, and then compute a homography matrix $\mathcal{H}_{s \rightarrow n}^i \in \mathbb{R}^{3\times3}$~\cite{zhou2019stnhomography} for each pair of $P_{s}^i$ and $P_{n}^i$, based on the four corresponding corner points of the two patches. Concretely, $\mathcal{H}_{s \rightarrow n}^i$ serves as a perspective transformation to relate the pixel coordinates in the two patches, formulated as:
\begin{equation}
\left[\begin{array}{c}
x_n^{i} \\
y_n^{i} \\
1
\end{array}\right]=\mathcal{H}_{s \rightarrow n}^i\left[\begin{array}{c}
x_s^{i} \\
y_s^{i} \\
1
\end{array}\right]=\left[\begin{array}{ccc}
h_{11}^i & h_{12}^i & h_{13}^i \\
h_{21}^i & h_{22}^i & h_{23}^i \\
h_{31}^i & h_{32}^i & h_{33}^i
\end{array}\right]\left[\begin{array}{c}
x_s^{i} \\
y_s^{i} \\
1
\end{array}\right]
\label{eq:homography}
\end{equation}
where $(x_n^{i},y_n^{i})$ and $(x_s^{i},y_s^{i})$ are the pixel coordinates in the normalized template patch and the unnormalized source patch, respectively. To compute the homography matrix $\mathcal{H}_{s \rightarrow n}^i$, we directly leverage the OpenCV API, which takes as inputs the corner points of the two patches and is implemented by using least-squares optimization and the Levenberg-Marquardt method~\cite{gavin2019levenberg}.
After obtaining $\mathcal{H}_{s \rightarrow n}^i$, we can transform the source patch $P_{s}^i$ to the normalized patch $P_{n}^i$ according to Eq.~\ref{eq:homography}.

Moreover, the normalized patches $P_{n}$ can further be transformed to target garment patches $P_{t}$ by utilizing the target pose $J_t$, which can be obtained from the target person $I_t$ via ~\cite{openpose}. The mechanism of that backward transformation is equivalent to the forward one in Eq.~\ref{eq:homography}, i.e., computing the homography matrix $\mathcal{H}_{n \rightarrow t}^i$ based on the four point pairs extracted from the normalized patch $P_{n}^i$ and the target pose $J_t$. The recovered target patches $P_{t}$ can then be stitched to form the warped garment $G_t$ that will be sent to the texture synthesis branch in Fig.~\ref{fig:framework} to generate more realistic garment transfer results. We can also regard $\mathcal{H}_{s \rightarrow t} = \mathcal{H}_{n \rightarrow t} \cdot \mathcal{H}_{s \rightarrow n}$ as the combined deformation matrix that warps the source garment to the target person pose, bridged by an intermediate normalized patch representation that is helpful for disentangling garment styles and spatial features.

\subsection{Dual-path Conditional StyleGAN2}~\label{tab:sec3-2}
Motivated by the impressive performance of StyleGAN2~\cite{karras2020stylegan2} in the field of image synthesis, our PASTA-GAN++ inherits the main architecture of StyleGAN2 and modifies it to a conditional version (see Fig.~\ref{fig:framework}). In the synthesis network, the normalized patches $P_n$ are projected to the style code $w$ through a style encoder followed by a mapping network, which is spatial-agnostic benefiting from the patch-routed disentanglement module. In parallel, the target pose $J_t$ is transferred into a pose feature map $f_{pose}$ by the pose encoder. Besides, to directly copy the protected regions (i.e., head, hand, and foot) from the target person to the synthesized result and guarantee the consistence of skin color between the target person and the synthesized result, an identity encoder which takes as inputs the protected region $H_t$ and the median skin color map $C_t$ is introduced to extract multi-scale feature maps of the identity information. Thereafter, the synthesis network takes as input the pose feature map and leverages the style code as the injected vector for each synthesis block to generate the try-on result $\tilde{I_t}'$. Besides, the multi-scale feature maps are separately concatenated with the corresponding feature maps in the synthesis blocks.

However, the standalone conditional StyleGAN2 is insufficient to generate compelling garment details especially in the presence of complex textures or logos. For example, although the illustrated $\tilde{I_t}'$ in Fig.~\ref{fig:framework}(b) can recover accurate garment style (color and shape) given the spatial-agnostic style code $w$, it lacks the complete texture pattern. The reasons for this are twofold: First, the style encoder projects the normalized patches into a one-dimensional vector, resulting in loss of high frequency information. Second, due to the large variety of garment texture, learning the local distribution of the particular garment details is highly challenging for the basic synthesis network.

To generate more accurate garment details, instead of only having a one-way synthesis network, we intentionally introduce a dual-path conditional StyleGAN2 which splits the synthesis network into two branches after the $256 \times 256$ synthesis block, namely the Style Synthesis Branch (SSB) and the Texture Synthesis Branch (TSB). 
The SSB modifies the last synthesis block of StyleGAN2 to a patch-guided parsing synthesis block, which aims to generate intermediate try-on result $\tilde{I_t}'$ with accurate garment style and predicts a human parsing $S_t$ that indicates the precise shape of different body parts (i.e., garment, arm, leg) complying with the target person pose.
It is worth noting that, compared with predicting several one-hot masks for various garment categories, as is done in our previous PASTA-GAN~\cite{xie2021pastagan}, predicting a unified human parsing can avoid ambiguity of garment labels. Specifically, in our previous PASTA-GAN, there could be overlaps between the predicted upper mask and lower mask, resulting in category confusion in the try-on result. However, since the human parsing can only indicate one label for each region, the category confusion issue can be completely eliminated.
Moreover, we experimentally found that predicting several garment masks for the various garment categories makes the network difficult to converge, whereas predicting a unified human parsing stabilizes the network training.

The purpose of TSB is to exploit the warped garment $G_t$, which has rich texture information to guide the synthesis path, and generate high-quality try-on results. 
We introduce a novel spatially-adaptive residual module specifically before the final synthesis block of the TSB, to embed the warped garment feature $f_g$ into the intermediate features of the synthesis network, which is beneficial for successfully synthesizing texture of the final try-on result $I_t'$. The detail of this module will be described in the following section.

\begin{figure}[t]
\centering
\includegraphics[width=\linewidth]{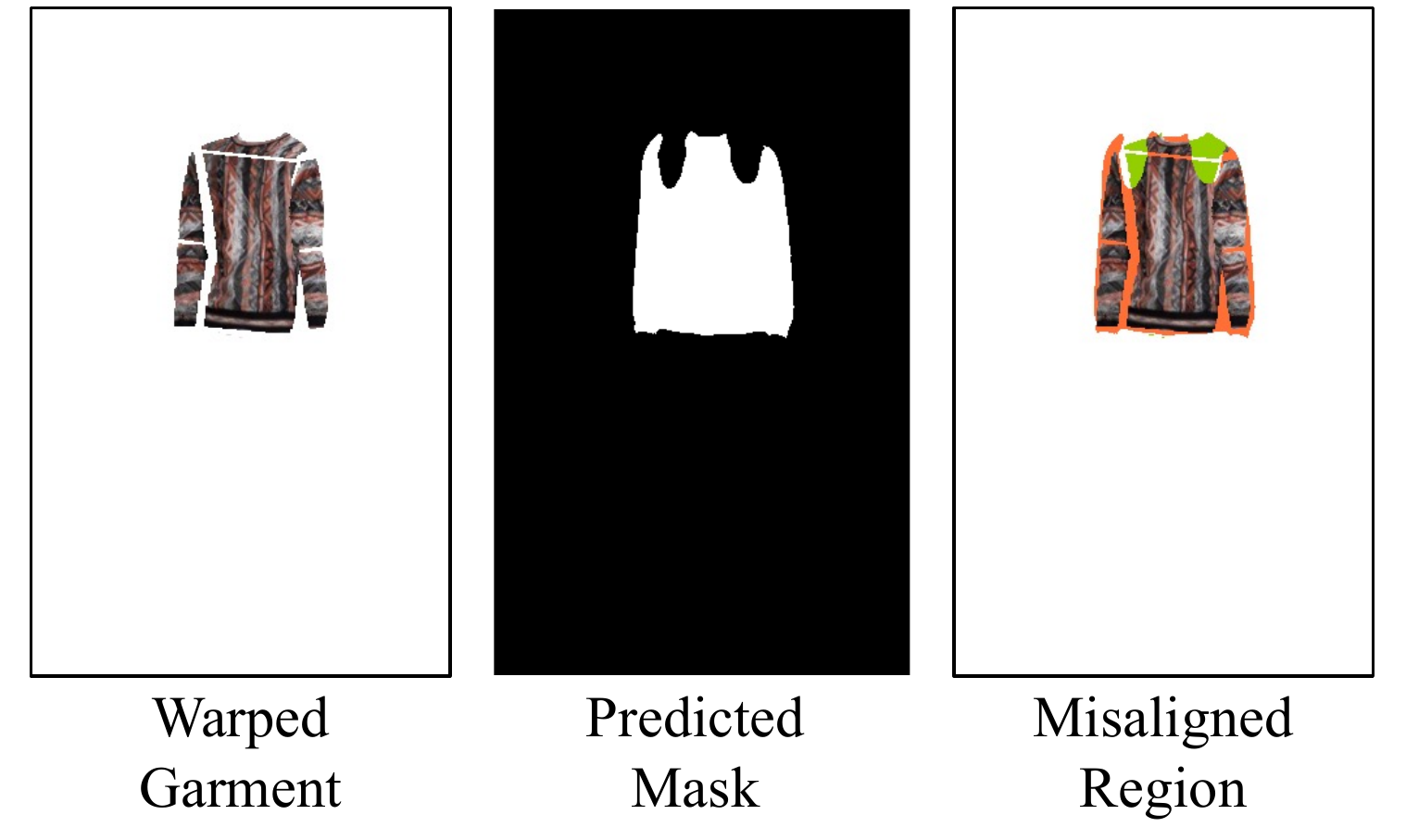}
\caption{Illustration of misalignment between the warped garment and the predicted mask , in which the warped garment is directly obtained by stitching the target patches together while the predicted mask is derived from the predicted human parsing $S_t$. The orange and green region represent the region to be inpainted and to be removed, respectively.\vspace{-10pt}}
\label{fig:misalignment}
\end{figure}

\begin{figure}[t]
  \centering
  \includegraphics[width=1.0\hsize]{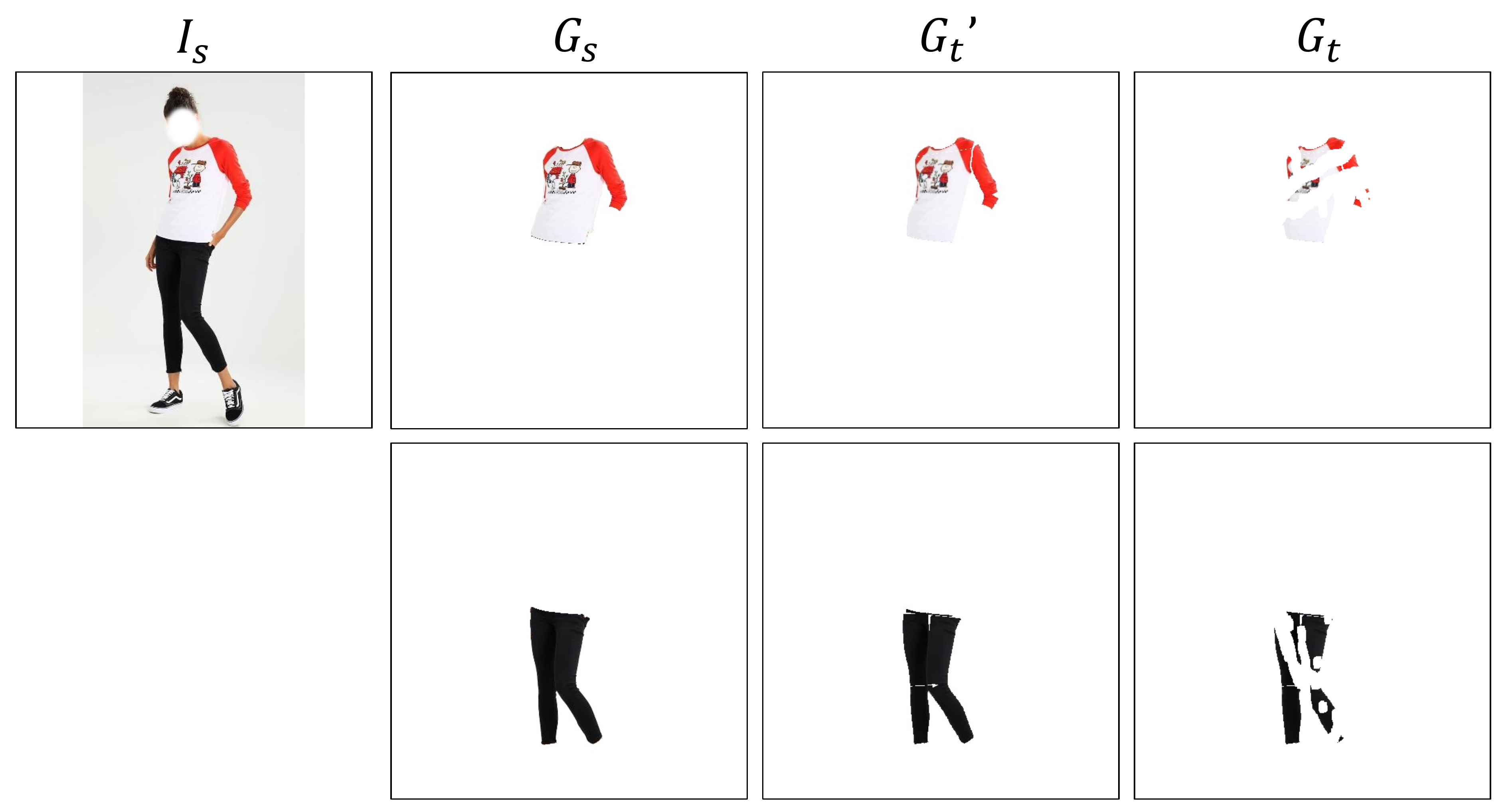}
  \caption{Comparison among the source garment and different warped garments, where $G_t'$ and $G_t$ are obtained by directly stitching the warped patches together and by applying a random mask to erase parts of $G_t'$, respectively. Please zoom in for more details.}
  \vspace{-3mm}
  \label{fig:warped garment}
\end{figure}

\subsection{Spatially-adaptive Residual Module}~\label{tab:sec:3-3}
Given the style code that factors out the spatial information and only keeps the style information of the garment, the style synthesis branch in Fig.~\ref{fig:framework} can accurately predict the mean color and the shape of the target garment. However, its inability to model the complex texture raises the need to exploit the warped garment $G_t$ to provide features that encode high-frequency texture patterns, which is in fact the motivation of the target garment reconstruction in Fig.~\ref{fig:patch-routed}. 

However, as the coarse warped garment $G_t$ is directly obtained by stitching the target patches together, its shape is inaccurate and usually misaligned with the garment masks $M_g$ (including the upper and the lower masks), in which the garment masks are obtained from the predicted human parsing $S_t$. Fig.\ref{fig:misalignment} takes the upper garment as example and explicitly shows the misalignment. Such shape misalignment in $G_t$ will consequently reduce the quality of the extracted warped garment feature $f_g$. 

To address this issue, we introduce the spatially-adaptive residual module between the last two synthesis blocks in the texture synthesis branch as shown in Fig.~\ref{fig:framework}. This module is comprised of a garment encoder and two spatially-adaptive residual blocks with feature inpainting mechanism to modulate intermediate features by leveraging the inpainted warped garment features. 

To be specific on the feature inpainting process, we first remove the part of $G_t$ that falls outside of $M_g$ (green region in Fig.\ref{fig:misalignment}), and explicitly inpaint the misaligned regions of the feature map within $M_g$ with average feature values (orange region in Fig.~\ref{fig:misalignment}). The inpainted feature map can then help the final synthesis block infer reasonable texture in the misaligned parts. 

Therefore given the predicted garment mask $M_g$, the coarse warped garment $G_{t}$, and its mask $M_t$, the process of feature inpainting can be formulated as:
\begin{equation}
M_{align} = M_g \cap M_{t},
\end{equation}
\begin{equation}
M_{misalign} = M_g - M_{align},
\end{equation}
\begin{equation}
f'_{g} = \mathcal{E}_g(G_{t} \odot M_g),
\end{equation}
\begin{equation}
f_{g} = f_{g}' \odot (1-M_{misalign}) + \mathcal{A}(f_{g}'\odot M_{align}) \odot M_{misalign},
\end{equation}
where $\mathcal{E}_g(\cdot)$ represents the garment encoder and $f_{g}'$ denotes the raw feature map of $G_t$ masked by $M_g$. $\mathcal{A}(\cdot)$ calculates the average garment features and $f_g$ is the final inpainted feature map.

Subsequently, inspired by the SPADE ResBlk from SPADE~\cite{park2019spade}, the inpainted garment features are used to calculate a set of affine transformation parameters that efficiently modulate the normalized feature map within each spatially-adaptive residual block. The normalization and modulation process for a particular sample $h_{z, y, x}$ at location ($z \in C, y \in H, x \in W$) in a feature map can then be formulated as:
\begin{equation}
\gamma_{z, y, x}(f_g) \frac{h_{z,y,x}-\mu_{z}}{\sigma_{z}}+\beta_{z, y, x}(f_g),\label{eq:modulate}
\end{equation}
where $\mu_{z}=\frac{1}{H W} \sum_{y, x} h_{z, y, x}$ and $\sigma_{z}=\sqrt{\frac{1}{H W} \sum_{y, x}\left(h_{z, y, x} - \mu_{z}\right)^{2}}$ are the mean and standard deviation of the feature map along channel $C$. $\gamma_{z, y, x}(\cdot)$ and $\beta_{z, y, x}(\cdot)$ are the convolution operations that convert the inpainted features to affine parameters.

Eq.~\ref{eq:modulate} serves as a learnable normalization layer for the spatially-adaptive residual block to better capture the statistical information of the garment feature map, thus helping the synthesis network to generate more realistic garment texture. 

With the modulated intermediate feature maps produced by the spatially-adaptive residual module, the texture synthesis branch can effectively utilize the reconstructed warped garment and generate the final compelling try-on result with high-frequency texture patterns.


\subsection{Loss Functions and Training Details}~\label{tab:sec3-4}
As paired training data is unavailable, our PASTA-GAN++ is trained unsupervised via image reconstruction. During training, we utilize the reconstruction loss $\mathcal{L}_{rec}$ and the perceptual loss~\cite{johnson2016perceptual} $\mathcal{L}_{perc}$ for both the coarse try-on result $\widetilde{I}'$ and the final try-on result $I'$:
\begin{equation}
\mathcal{L}_{rec} = \sum_{I \in \{\widetilde{I}', I'\}}\|I - I_s \|_1,
\end{equation}

\begin{equation}
\mathcal{L}_{perc} = \sum_{I \in \{\widetilde{I}', I'\}}\sum_{k=1}^{5} \lambda_{k}\left\|\phi_{k}(I)-\phi_{k}\left(I_s \right)\right\|_{1},
\end{equation}
where $\phi_k(I)$ denotes the $k$-th feature map in a VGG-19 network~\cite{DBLP:journals/corr/SimonyanZ14a} pre-trained on the ImageNet~\cite{ILSVRC15} dataset.
We also use the pixel-wise cross-entropy loss $L_{CE}$ between the predicted human parsing $S_t$ and its ground truth $S$, which is obtained via~\cite{Gong2019Graphonomy}.
Besides, we separately calculate the adversarial loss $\mathcal{L}_{GAN}$ for $S_t$, $\widetilde{I'}$, and $I'$, which is the same as in StyleGAN2~\cite{karras2020stylegan2}. The total loss can be formulated as:
\begin{equation}
\mathcal{L} = \mathcal{L}_{GAN} + \lambda_{rec}\mathcal{L}_{rec} + \lambda_{perc}\mathcal{L}_{perc} + \lambda_{ce}\mathcal{L}_{CE},
\end{equation}
where $\lambda_{rec}$, $\lambda_{perc}$, and $\lambda_{ce}$ are trade-off hyper-parameters.

Note, during training, although the source and target pose are the same, the coarse warped garment $G_t$ is not identical to the intact source garment $G_s$, due to the crop mechanism in the patch-routed disentanglement module. More specifically, the quadrilateral crop for $G_s$ is by design not seamless/perfect and there will accordingly often exist some small seams between adjacent patches in $G_t$ as well as incompleteness along the boundary of the torso region. To further reduce the training-test gap of the warped garment, we use the random mask from~\cite{liu2018partialconv} to erase parts of the warped garment with a probability of $\alpha$. 
Such an erasing operation can imitate self-occlusion in the source person image. Fig.~\ref{fig:warped garment} illustrates the process by displaying the source garment $G_s$, the warped garment $G_t'$ that is obtained by directly stitching the warped patches together, and the warped garment $G_t$ that is sent to the network. We can observe a considerable difference between $G_s$ and $G_t$. 

\begin{table*}[t]
    \caption{The the Fr$\mathbf{\acute{e}}$chet Inception Distance score (FID)~\cite{parmar2021cleanfid} and Human Evaluation score (HE) among different methods under various try-on settings on the high-resolution version of the UPT ($512\times320$) dataset.}
    \def\arraystretch{0.9}
    \small
    \tabcolsep 13pt
    \centering
    \begin{tabular}{l|cc|cc|cc|cc}
        \toprule
        \multirow{2}*{Method} & \multicolumn{2}{c|}{Upper-body}  & \multicolumn{2}{c|}{Lower-body} & \multicolumn{2}{c|}{Full-body} & \multicolumn{2}{c}{Dress} \\
        \cmidrule{2-9}
        & FID $\downarrow$ & HE $\uparrow$ & FID $\downarrow$ & HE $\uparrow$ & FID $\downarrow$ & HE $\uparrow$ & FID $\downarrow$ & HE $\uparrow$ \\
        \midrule
        PB-AFN~\cite{ge2021pfafn}     &  11.79    &  19.72\%    &  -      &  -      &  -      &  -      &  -      &  -   \\
        DiOr~\cite{Cui2021dior}       &  25.87    &  0.83\%    &  28.17    &  1.82\%    &  26.24    &  5.87\%    &  54.11    &  0.51\% \\
        wFlow~\cite{dong2022wflow}    &  14.30  &  8.33\%    &  14.71  &  15.76\%    &  16.36  &  18.10\%    &  34.81    &  15.64\% \\
        PWS~\cite{albahar2021pose}    &  11.15  &  10.83\%    &  13.29  &  6.37\%    &  12.60  &   16.19\%    &  28.44    &  9.23\% \\
        \midrule
        PASTA-GAN (Ours)~\cite{xie2021pastagan}                          &  6.69    &  13.33\%    &  8.65  & 18.79\%     &  9.51  & 19.37\%     &  32.03  &  7.69\% \\
        PASTA-GAN++ (Ours)                          &  \textbf{5.65}    &  \textbf{46.94\%}    &  \textbf{7.40}  & \textbf{57.27\%}     &  \textbf{8.20}  & \textbf{40.48\%}     &  \textbf{22.89}  & \textbf{66.92\%} \\
        \bottomrule
    \end{tabular}
    \label{tab:unpaired quantitative result}
\end{table*}

\begin{figure*}[t]
\centering
\includegraphics[width=\linewidth]{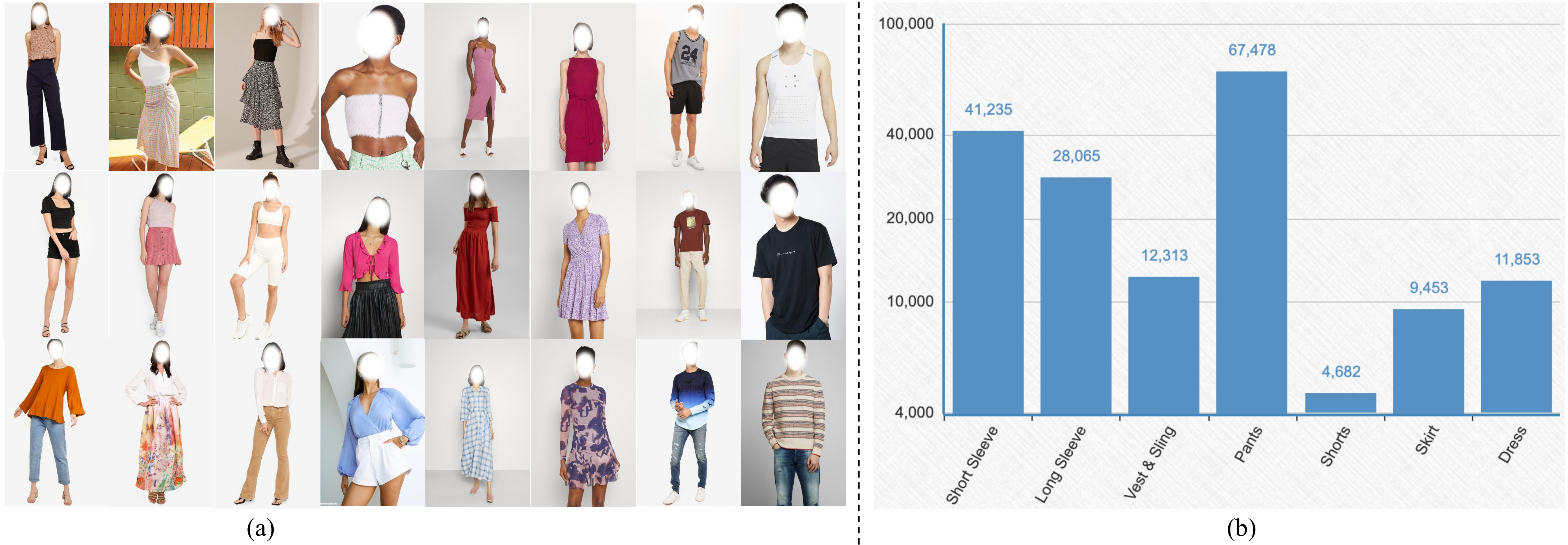}
\caption{(a) Examples of our newly collected UPT dataset. (b) The category distribution of our UPT dataset. Please zoom in for more details. \vspace{-10pt}}
 \vspace{-3mm}
\label{fig:dataset}
\end{figure*}

\section{Experiments}
\subsection{Experiment Setup}
\noindent\textbf{Datasets.} 
To explore the proposed high-resolution unpaired virtual try-on algorithm, we construct a high-quality fashion model dataset, named the UnPaired virtual Try-on (UPT) dataset, which is composed of newly collected front-view fashion model images from a set of e-commerce websites and the front-view subsets of two existing virtual try-on benchmark datasets (the MPV~\cite{dong2019mgvton} dataset and the DeepFashion~\cite{liuLQWTcvpr16DeepFashion} dataset).
Specifically, the UPT dataset contains 102,468 high-resolution ($512 \times 320$), half- and full-body front-view images of persons wearing a large variety of garments, e.g., long/short sleeve shirt, vest, sling, pants, shorts, skirts, dresses, etc. UPT is further split into a training set of 93,452 images and a testing set of 9,016 images. Fig. ~\ref{fig:dataset} displays some examples of UPT and the distribution of various garment types of UPT, demonstrating the garment diversity of the UPT dataset which is crucial for the unpaired virtual try-on task. Personally identifiable information (i.e. face information) has been masked out.

\noindent\textbf{Metrics.} We apply the Fr$\mathbf{\acute{e}}$chet Inception Distance (FID)~\cite{parmar2021cleanfid} to measure the similarity between real and synthesized images, and perform a human evaluation study to quantitatively evaluate the synthesis quality of different methods. For the human evaluation, we design four questionnaires corresponding to the four types of virtual try-on settings (i.e., upper-, lower-, full-body, and dresses virtual try-on). In each questionnaire, we randomly select 30 try-on results generated by our PASTA-GAN++ and the other compared methods. Then, for each questionnaire, we invite 30 volunteers to complete the 30 tasks by choosing the most realistic try-on results given the person images and the garment images. Finally, the human evaluation score is calculated as the chosen percentage for a particular method. 

\noindent\textbf{Implementation Details.} Our PASTA-GAN++ is implemented by using PyTorch~\cite{paszke2019pytorch} and is trained on 8 Tesla V100 GPUs. During training, the batch size is set to 24 and the model is trained for 8 million iterations with a learning rate of 0.0005 using the Adam optimizer~\cite{Kingma2014adam} with $\beta_1=0$ and $\beta_2=0.99$. The loss hyper-parameters $\lambda_{rec}$, $\lambda_{perc}$, and $\lambda_{ce}$ are set to 10, 20, and 30, respectively. The hyper-parameters for the random erasing probability $\alpha$ is set to 0.9. The main network architecture (e.g., style encoder, identity encoder, etc.) follows our PASTA-GAN~\cite{xie2021pastagan}. 

\noindent\textbf{Baselines.} 
To validate the effectiveness of our PASTA-GAN++, we compare it with the state-of-the-art high-resolution methods that have released the official code and pre-trained weights. This includes one garment-to-person virtual try-on method PB-AFN~\cite{ge2021pfafn}, and four person-to-person virtual try-on methods DiOr~\cite{Cui2021dior}, wFlow~\cite{dong2022wflow}, PWS~\cite{albahar2021pose}, and PASTA-GAN~\cite{xie2021pastagan}.\footnote{For all these prior approaches, research use is permitted according to the respective licenses. Note, we are unable to compare with~\cite{sarkar2021posestylegan},~\cite{ira2021vogue} and~\cite{neuberger2020oviton} as they have not released their code or pre-trained model.} 
To obtain high-resolution virtual try-on results for PB-AFN, we re-train it using the high-resolution version of the MPV ($512 \times 320$) dataset~\cite{dong2019mgvton}.
For Dior and PWS, we use the pre-trained models trained on the high-resolution version of the Deepfashion~\cite{liuLQWTcvpr16DeepFashion} ($512 \times 384$) dataset. For wFlow, we use its pre-trained model, which is trained on the Dancn50k~\cite{dong2022wflow} and Deepfashion datasets. 
For our previous PASTA-GAN~\cite{xie2021pastagan}, we retain its overall architecture, adjust it to accept high-resolution inputs and then re-train the network on the high-resolution version of the UPT dataset. 
During testing, the person-to-person methods~\cite{Cui2021dior,dong2022wflow,albahar2021pose,xie2021pastagan} directly take as inputs two person images and synthesize the try-on result, while for the garment-to-person method~\cite{ge2021pfafn}, we first extract the desired garment from the person image and regard it as the in-shop garment to meet the input requirements of garment-to-person approaches.
Besides, for the person-to-person methods~\cite{Cui2021dior,dong2022wflow,albahar2021pose,xie2021pastagan}, we conduct experiments under four try-on settings (i.e., upper-, lower-, full-body, and dresses try-on.) to extensively compare the performance of the different methods  under various settings, while for the garment-to-person method PB-AFN~\cite{ge2021pfafn}, we only re-train one model for the upper-body virtual try-on since garment-person pairs for lower garments and dresses do not exist in the MPV dataset.

\subsection{Quantitative Comparison with the state-of-the-art methods}\label{sec:4.1}
As reported in Table~\ref{tab:unpaired quantitative result}, our proposed PASTA-GAN++ outperforms the baseline methods under all four testing setting, namely, the upper-, lower-, full-body, and dresses try-on settings, illustrating that PASTA-GAN++ is capable of generating photo-realistic virtual try-on results for arbitrary garment categories. 
Besides, PASTA-GAN++ surpasses our previous PASTA-GAN~\cite{xie2021pastagan} by a large margin under the dresses try-on setting, demonstrating the efficiency of the more dedicated garment patches and the introduction of the patch-guided parsing synthesis block.
It is worth to note that, for the lower-body, full-body, and dresses try-on, due to the large diversity of the lower garments, all methods are prone to get higher FID scores.

\begin{figure*}[t]
\centering
\includegraphics[width=\linewidth]{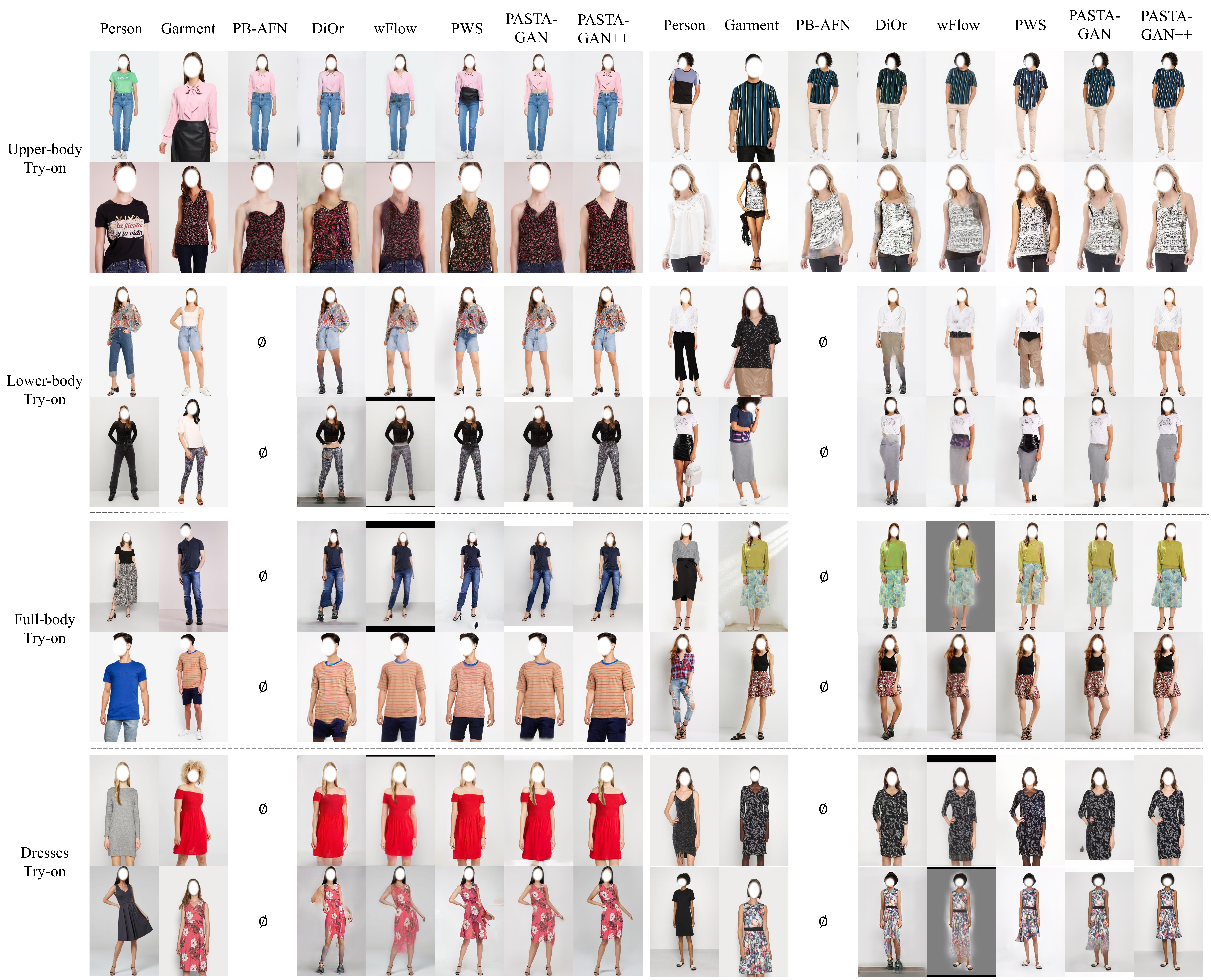}
\caption{Qualitative comparisons between our PASTA-GAN++ and the existing state-of-the-art methods under four try-on setting (i.e., upper-, lower-, full-body, and dresses try-on). Note, PB-AFN is only evaluated on the upper-body try-on task due to the lack of dress and lower-body garment-person pairs. Please Zoom in for more details.\vspace{-10pt}}
\label{fig:comp_512}
\end{figure*}

\begin{figure*}[t]
  \centering
  \includegraphics[width=1.0\hsize]{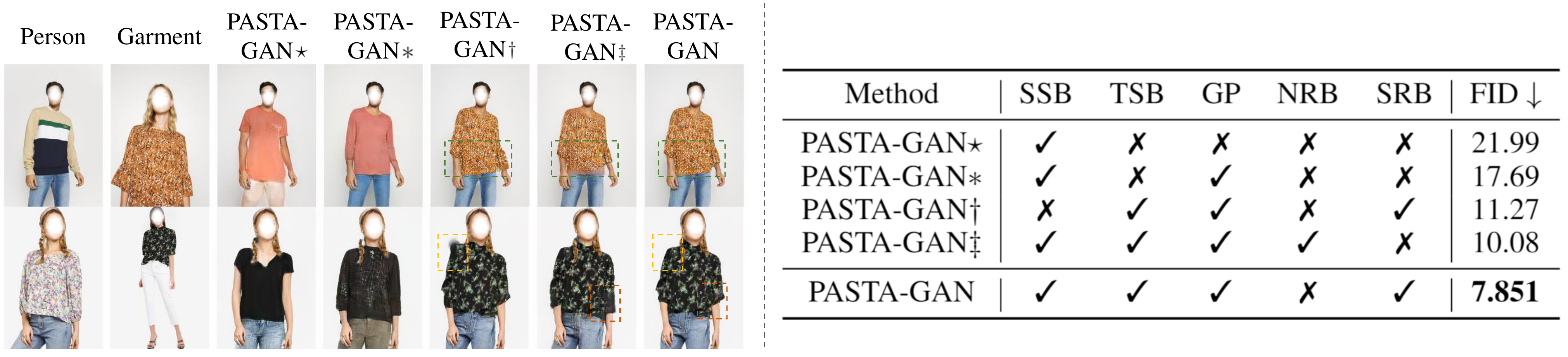}
  \caption{Qualitative results and quantitative results of the ablation study with different configurations, in which SSB, TSB, GP, NRB, SRB refer to the style synthesis branch, texture synthesis branch, garment patches, normal residual blocks, and spatially-adaptive residual blocks, respectively.}
  \vspace{-2mm}
  \label{fig:ablation_study}
\end{figure*}


\subsection{Qualitative Comparison with the state-of-the-art methods}\label{sec:4.2}
Fig.~\ref{fig:comp_512} shows a comprehensive visual comparison among different virtual try-on methods, from which we can conclude the superiority of our PASTA-GAN++ in two-folds. 
First, PASTA-GAN++ better preserves the garment texture.
PB-AFN~\cite{ge2021pfafn} and DiOr~\cite{Cui2021dior} directly employ the pixel-level appearance flow to conduct texture transfer. However, due to the lack of supervision in form of a ground truth flow, the estimated pixel-level appearance flow is prone to imprecisions, leading to distorted texture transfer results (e.g., the results of PB-AFN and DiOr in the second row of the upper-body transfer in Fig.~\ref{fig:comp_512}). wFlow~\cite{dong2022wflow} collaboratively combines a 2D pixel-level and a 3D vertex-level appearance flow to obtain more precise 
warped garments. However, wFlow tends to synthesize blurred results since it uses the predicted fusion mask to combine two warped garments from two different appearance flows, which will average the pixel values and is detrimental for the visual quality (e.g., the results of wFlow in the upper-body and lower-body transfer in Fig.~\ref{fig:comp_512}). PWS~\cite{albahar2021pose} uses the appearance flow derived from the Densepose~\cite{guler2018densepose}. However, it uses the pre-defined segmentation from Densepose to separate bodies into different parts, which may be inconsistent with the real human parsing, leading to incorrect garment transfer in some particular scenarios (e.g., the top-left result of PWS in the upper-body transfer and the bottom-right result of PWS in the lower-body transfer setting in Fig.~\ref{fig:comp_512}). Different from the baseline methods, due to the learning-free warping mechanism in our patch-routed disentanglement module and the feature inpainting process in the spatially-adaptive residual module, our PASTA-GAN++ can preserve most of the garment texture and refine the coarse warped result, resulting in more photo-realistic warped results.

Second, PASTA-GAN++ is capable of generating more precise garment shape. Existing baselines~\cite{ge2021pfafn,Cui2021dior,dong2022wflow,albahar2021pose} take the intact garment as inputs and extract the global garment feature for the try-on network, which is not sensitive enough to capture local details. Instead, PASTA-GAN++ utilizes the patch-based garment representation, which facilitates the network to encode fine-grain garment characteristics (e.g., the sleeve length, the collar shape, etc.) and generate garment shape with correct details. For example, as shown in the first column of the upper-body transfer result in Fig.~\ref{fig:comp_512}, PASTA-GAN++ can generate precise sleeve length and collar shape while other methods are prone to obtain artifacts in these region. On the other hand, compared with the previous PASTA-GAN~\cite{xie2021pastagan}, our proposed PASTA-GAN++ can generate more realistic garment shape around the upper-lower boundary (e.g., the results in the second column of the upper-body transfer in Fig.~\ref{fig:comp_512}). Besides, PASTA-GAN++ is able to generate more realistic dresses try-on results while PASTA-GAN usually generates artifacts around the lower part (e.g., the results in the dresses transfer in Fig.~\ref{fig:comp_512}).



\subsection{Ablation Studies}
In this section, we conduct ablation experiments to separately validate the effectiveness of the patch-routed disentanglement module and the spatially-adaptive residual module. Note, as for the dedicated patch representation and the patch-guided parsing synthesis blocks, we have validated their effectiveness in the quantitative and qualitative comparisons between PASTA-GAN++ and the previous PASTA-GAN~\cite{xie2021pastagan} in Sec.~\ref{sec:4.1} and Sec.~\ref{sec:4.2}.

\noindent\textbf{Patch-routed Disentanglement Module:} To validate its effectiveness, we train two PASTA-GANs without texture synthesis branch, denoted as PASTA-GAN$\star$ and PASTA-GAN$*$, which take the intact garment and the garment patches as input of the style encoder, respectively. 
As shown in Fig.~\ref{fig:ablation_study}, PASTA-GAN$\star$ fails to generate accurate garment shape. In contrast, the PASTA-GAN$*$ which factors out spatial information of the garment, can focus more on the garment style information, leading to the accurate synthesis of the garment shape. However, without the texture synthesis branch, both of them are unable to synthesize the detailed garment texture. The models with the texture synthesis branch can preserve the garment texture well as illustrated in Fig~\ref{fig:ablation_study}.

\noindent\textbf{Spatially-adaptive Residual Module} To validate the effectiveness of this module, we further train two PASTA-GANs with texture synthesis branch, denoted as PASTA-GAN$\dagger$ and PASTA-GAN$\ddagger$, 
which excludes the style synthesis branch and replaces the spatially-adaptive residual blocks with normal residual blocks, respectively.
Without the support of the corresponding components, both PASTA-GAN$\dagger$ and PASTA-GAN$\ddagger$ fail to fix the garment misalignment problem, leading to artifacts outside the target shape and blurred texture synthesis results. The full PASTA-GAN instead can generate try-on results with precise garment shape and texture details.
The quantitative comparison results in Fig.~\ref{fig:ablation_study} further validate the effectiveness of our designed modules.

\begin{figure}[t]
\centering
\includegraphics[width=\linewidth]{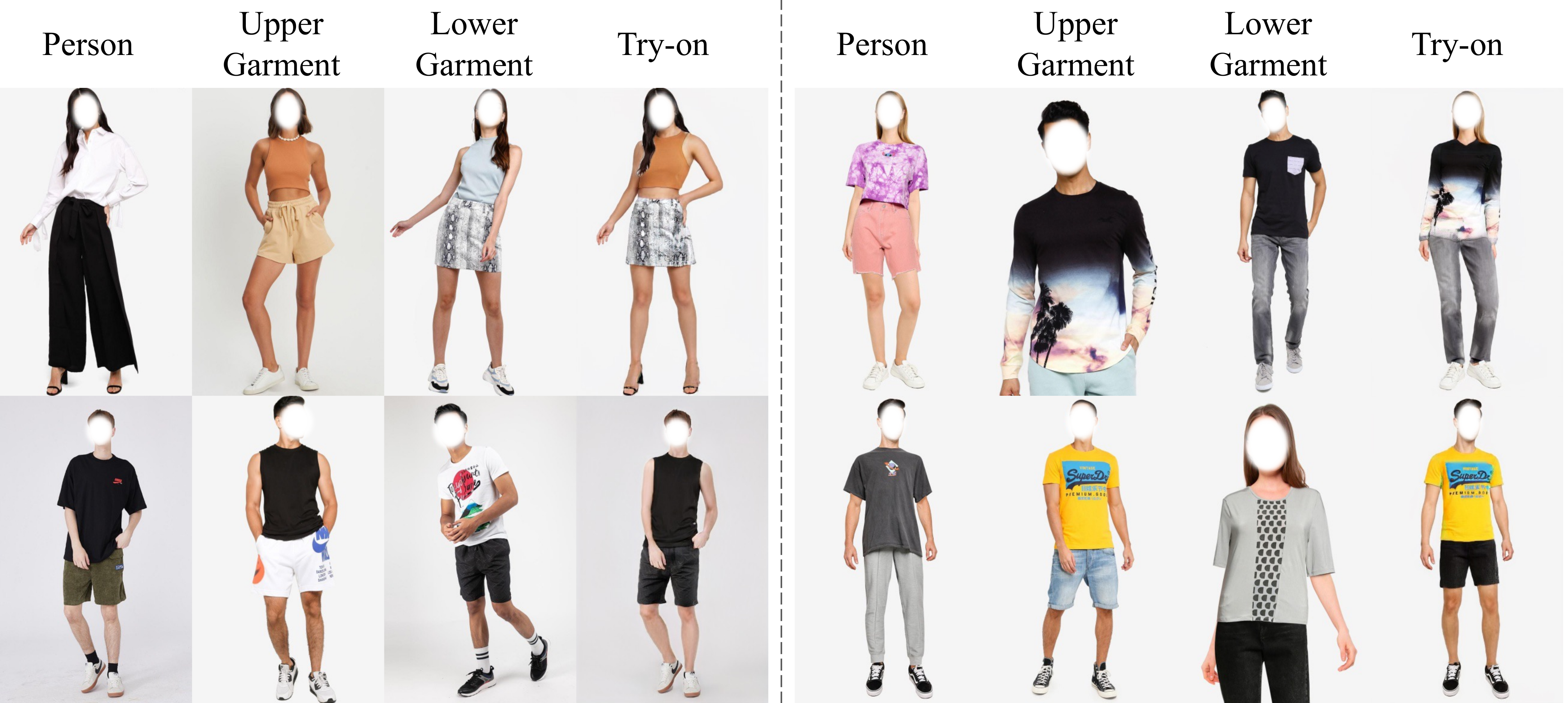}
\caption{Application: Outfit try-on. Our model can transfer outfits from different garment images into the specific person image within a single forward pass. Please Zoom in for more details.
\vspace{-10pt}
}
\label{fig:outfit_try-on}
\end{figure}

\begin{figure}[t]
\centering
\includegraphics[width=\linewidth]{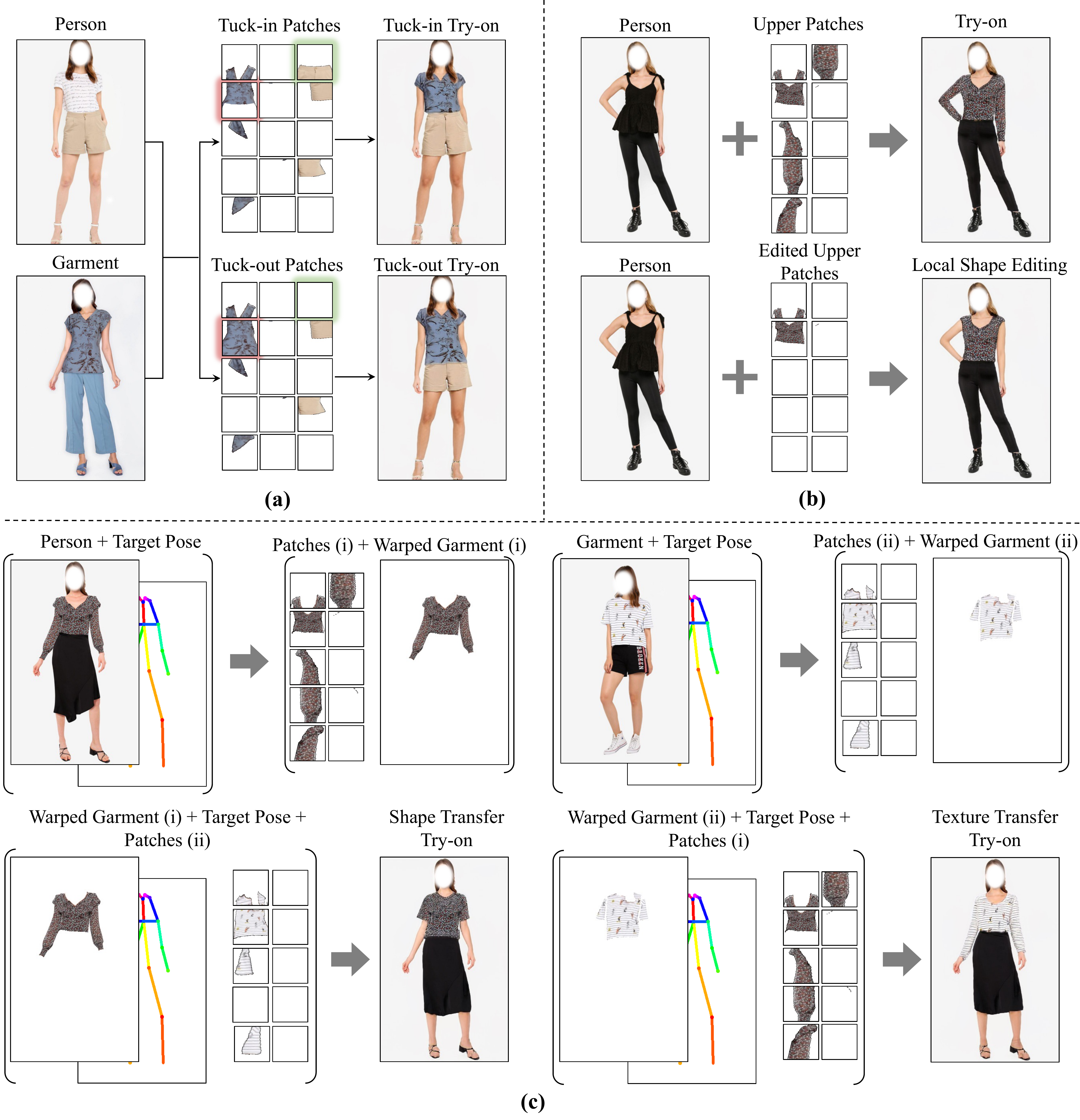}
\caption{Implementation mechanism of (a) dressing order control, (b) garment shape editing, and (c) garment shape/texture transfer. Please Zoom in for more details.}
\label{fig:apps-procedure}
 \vspace{-3mm}
\end{figure}

\begin{figure*}[t]
\centering
\includegraphics[width=\linewidth]{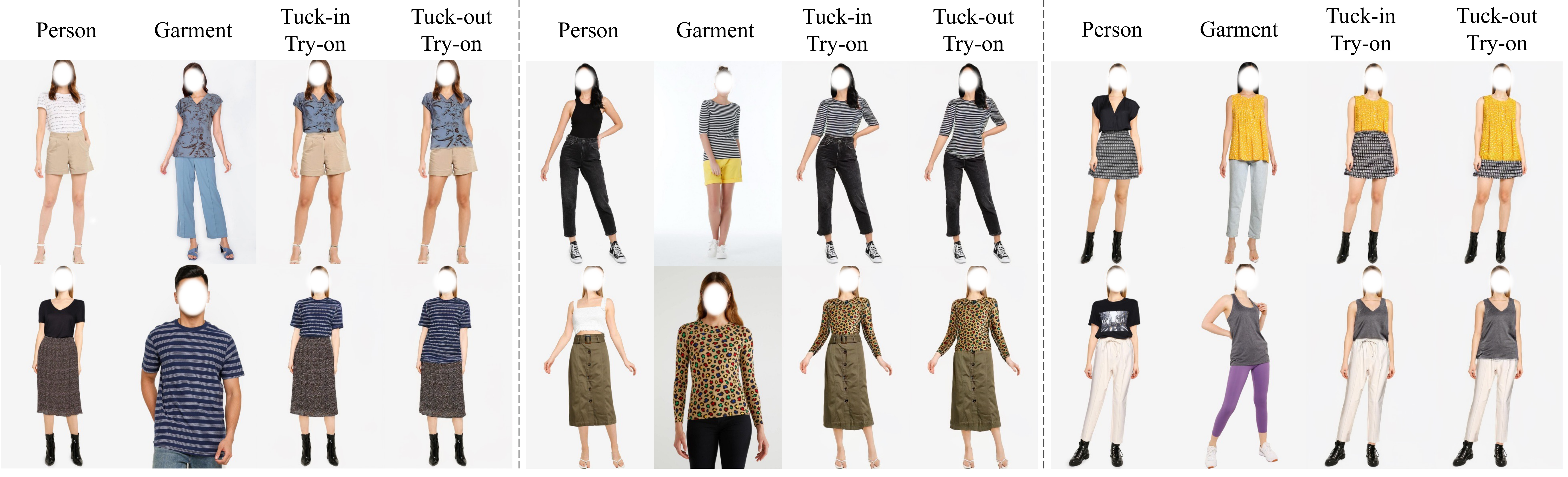}
\caption{Application: Dressing order control. Our model can adaptively tuck the upper garment into the lower or put it on the lower within the same pretrained model. Please Zoom in for more details.\vspace{-10pt}}
\label{fig:tuckin_tuckout}
\end{figure*}

\begin{figure*}[t]
\centering
\includegraphics[width=\linewidth]{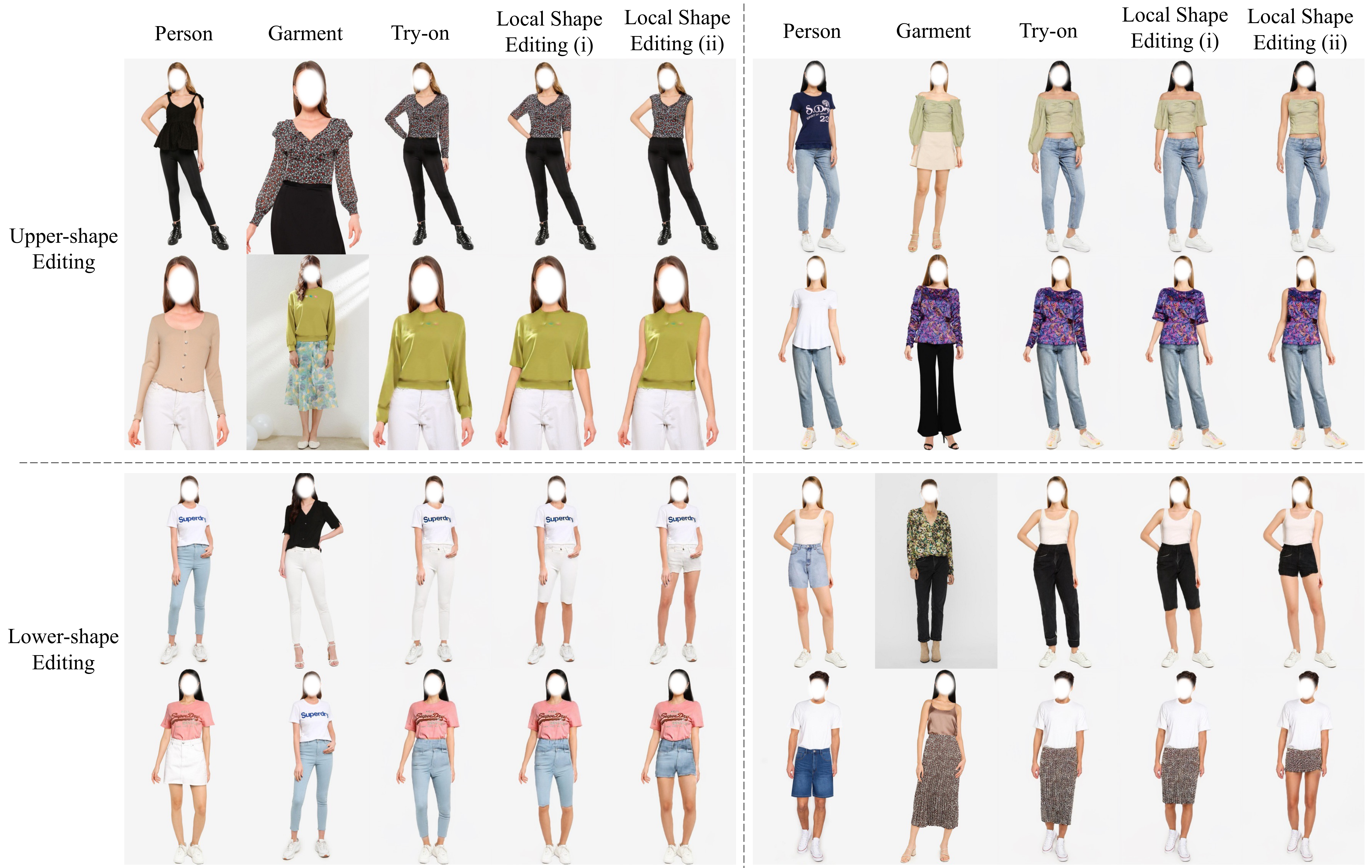}
\caption{Application: Local shape editing. Our model can edit the shape of the garment (e.g., the length of the sleeve, the length of the pants, etc.) by controlling the valid region in specific garment patches. Please Zoom in for more details.
\vspace{-10pt}}
\label{fig:garment_editing}
\end{figure*}

\begin{figure*}[t]
\centering
\includegraphics[width=\linewidth]{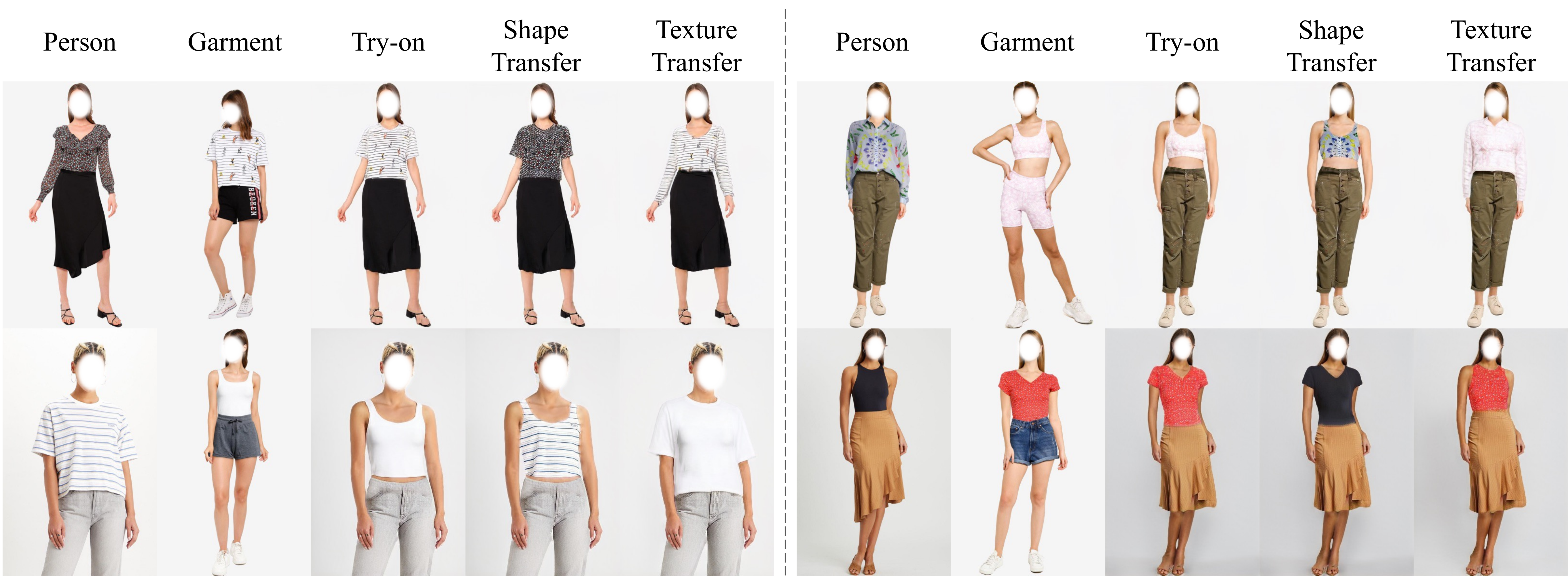}
\caption{Application: Reshape and texture transfer. 
Our model can achieve garment reshaping and texture transfer for a person image according to the shape and texture of a garment image.
Please Zoom in for more details.\vspace{-10pt}}
\label{fig:reshape_texturetransfer}
\end{figure*}

\section{Applications}
Due to the flexible patch-based garment representation, applications like garment editing can be easily conducted by directly editing some particular garment patches in the normalized space or combining garment patches and the warped garment from different garment images. In this section, we will introduce the applications PASTA-GAN++ supports.

\noindent\textbf{Outfits try-on.}
As shown in Fig.~\ref{fig:outfit_try-on}, our PASTA-GAN++ can integrate garment patches from different persons and transfer various garments onto a specified person to achieve outfit try-on within a single forward pass.

\noindent\textbf{Dressing order control.}
Our PASTA-GAN++ allows the user to control the dressing order of the upper and the lower garment to achieve tuck-in and tuck-out try-on (See Fig.~\ref{fig:tuckin_tuckout}). As shown in Fig.~\ref{fig:apps-procedure} (a), such an application can be implemented through editing the torso patch of the upper patches (i.e., patch with red shadow) and the torso patch of the lower patches (i.e., patch with green shadow).

\noindent\textbf{Local shape editing.} As shown in Fig.~\ref{fig:garment_editing}, our PASTA-GAN++ can change the local shape of a garment (e.g., sleeve shape, etc.) while preserving the remaining properties of the original garment (e.g., the shape of the neckline, etc.). Since each of the fine-grained garment patches only represents a specific local region of the garment, local shape editing can be achieved by changing some particular patches while retaining the other patches (Please refer to Fig.~\ref{fig:apps-procedure} (b) for an illustration of this process.).

\noindent\textbf{Garment shape/texture transfer.} 
Our PASTA-GAN++ is capable of integrating the attributes from different garments to synthesize a try-on result with a newly generated garment (i.e., the shape transfer try-on and the texture transfer try-on in Fig.~\ref{fig:reshape_texturetransfer}), in which the shape and the texture of the newly generated garment are derived from different images. 
Fig.~\ref{fig:apps-procedure} illustrates its implementation mechanism.
Specifically, given the person image and the garment image, PASTA-GAN++ first obtains the the normalized garment patches (i.e., patches (i) and (ii)) for each image and the warped garments (i.e., warped garment (i) and (ii)) that comply with the target pose, in which the normalized patches and the warped garment can separately provide the shape and the texture guidance for the synthesis network. 
Then, shape transfer try-on can be achieved by taking the warped garment (i) and the normalized patches (ii) as inputs to synthesize try-on results where garment shape and texture are derived from the garment image and person image, respectively, while texture transfer can be achieved by taking the warped garment (ii) and the normalized patches (i) as inputs to synthesize try-on result where the garment shape and texture are derived from the person image and garment image, respectively.

\section{Conclusion}
In this work, we take a step forwards to propose a versatile framework for high-resolution unpaired virtual try-on, named PAtch-routed SpaTially-Adaptive GAN++ (PASTA-GAN++), which supports unsupervised training with unpaired data, handling of arbitrary garment categories, and controllable garment editing. By utilizing the novel patch-routed disentanglement module and the spatially-adaptive residual module, PASTA-GAN++ effectively disentangles garment style and spatial information and generates realistic and accurate virtual-try on results without requiring auxiliary data or extensive online optimization procedures. Besides, thanks to the fine-grained patch-based garment representation and the introduction of the patch-guided parsing synthesis block into the dual-path StyleGAN2 generator, PASTA-GAN++ can handle arbitrary garment categories and support fine-grained garment editing during the try-on procedure. Extensive experiments on our high-resolution UnPaired Virtual Try-on (UPT) dataset highlight its superiority over existing virtual try-on methods and demonstrate its versatility for controllable garment editing. 
We believe that this work will inspire new versatile approaches to make full use of the large amount of available unlabeled data and fulfil various practical requirements in the real-world scenarios.

\ifCLASSOPTIONcompsoc
  \section*{Acknowledgments}
\else
  \section*{Acknowledgment}
\fi

This work was supported in part by the Shenzhen Fundamental Research Program (Project No.RCYX20200714114642083, No. JCYJ20190807154211365), the Guangdong Outstanding Youth Fund (Grant No. 2021B1515020061), the National Key R$\&$D Program of China under Grant (No. 2018AAA0100300), the Guangdong Province Basic and Applied Basic Research (Regional Joint Fund-Key) Grant (No. 2019B1515120039), and the National Natural Science Foundation of China under Grant (No. U19A2073) and (No. 61976233).

\ifCLASSOPTIONcaptionsoff
  \newpage
\fi



\bibliographystyle{IEEEtran}
\bibliography{PASTA-GAN}
%



%

\begin{IEEEbiography}[{\includegraphics[width=1in,height=1.25in,clip,keepaspectratio]{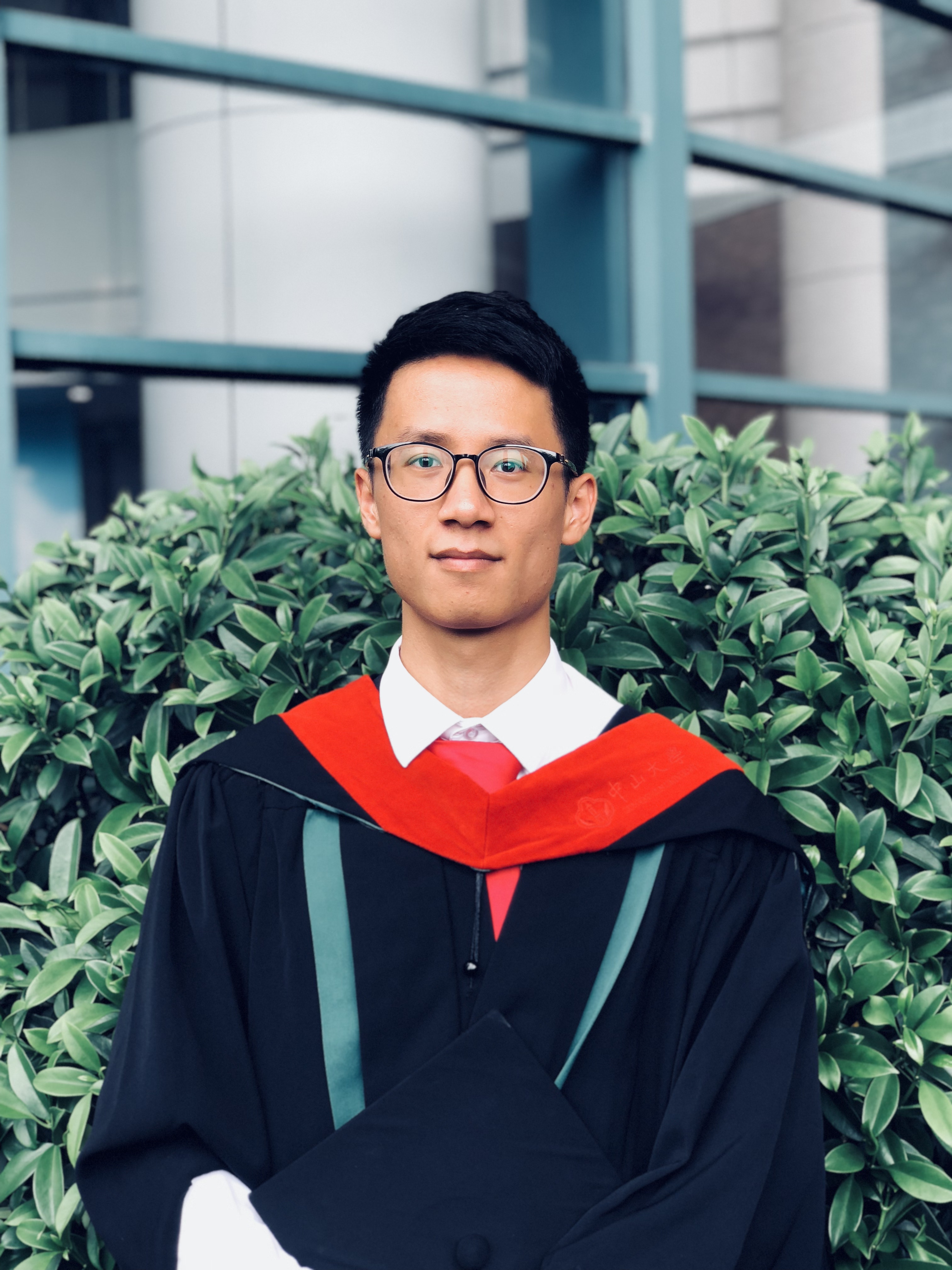}}]{Zhenyu Xie} received his BE and ME degree from Sun Yat-sen University, China. Now, he is working towards his PH.D. degree in Sun Yat-sen University. His research interests include human-centric image synthesis and controllable image manipulation.

\end{IEEEbiography}

\begin{IEEEbiography}[{\includegraphics[width=1in,height=1.25in,clip,keepaspectratio]{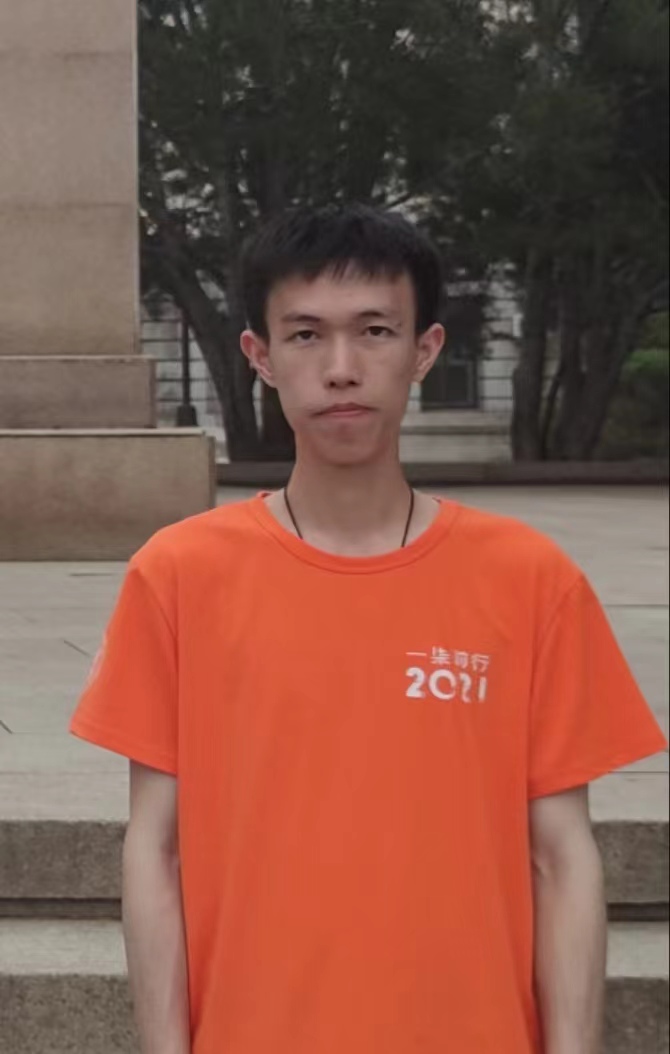}}]{Zaiyu Huang} received his BE degree from University of Science and Technology Beijing. He is currently pursing his ME degree in Sun Yat-sen University. His research interests include 2D/3D human-centric image synthesis.

\end{IEEEbiography}

\begin{IEEEbiography}[{\includegraphics[width=1in,height=1.25in,clip,keepaspectratio]{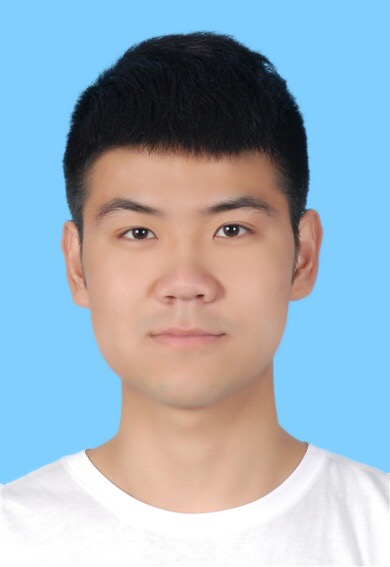}}]{Fuwei Zhao} received his BE degree and is currently pursing his ME degree from Sun Yat-sen University, China. His research interests includes 2D/3D human synthesis, 3D-aware human synthesis.

\end{IEEEbiography}

\begin{IEEEbiography}[{\includegraphics[width=1in,height=1.25in,clip,keepaspectratio]{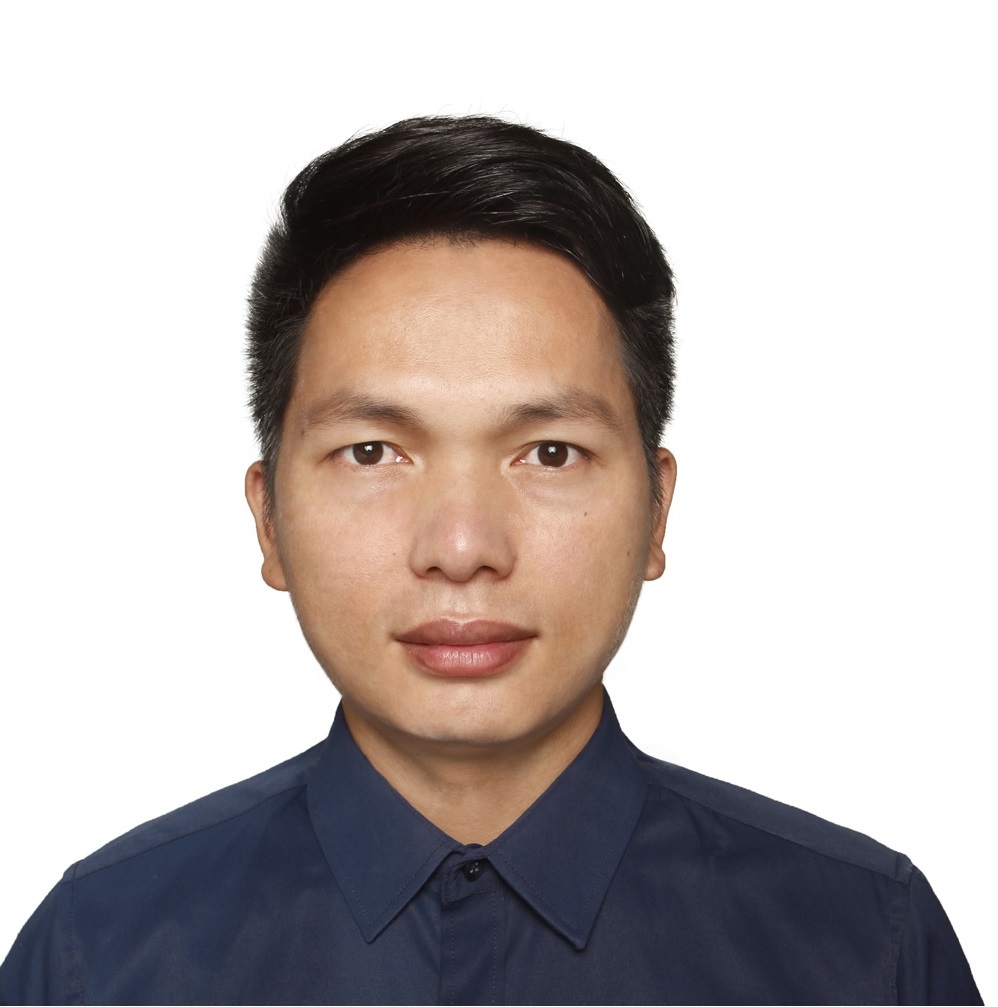}}]{Haoye Dong} is currently a postdoc researcher at Carnegie Mellon University. He received his Ph.D. degree from Sun Yat-sen University in 2019. His research mainly focuses on human-centric learning, enabling efficient 2D/3D avatar animation and properties editing intelligently. For example, virtual try-on, pose transfer, real-time digital human animation,  motion retargeting, and motion prediction. 

\end{IEEEbiography}

\begin{IEEEbiography}[{\includegraphics[width=1in,height=1.25in,clip,keepaspectratio]{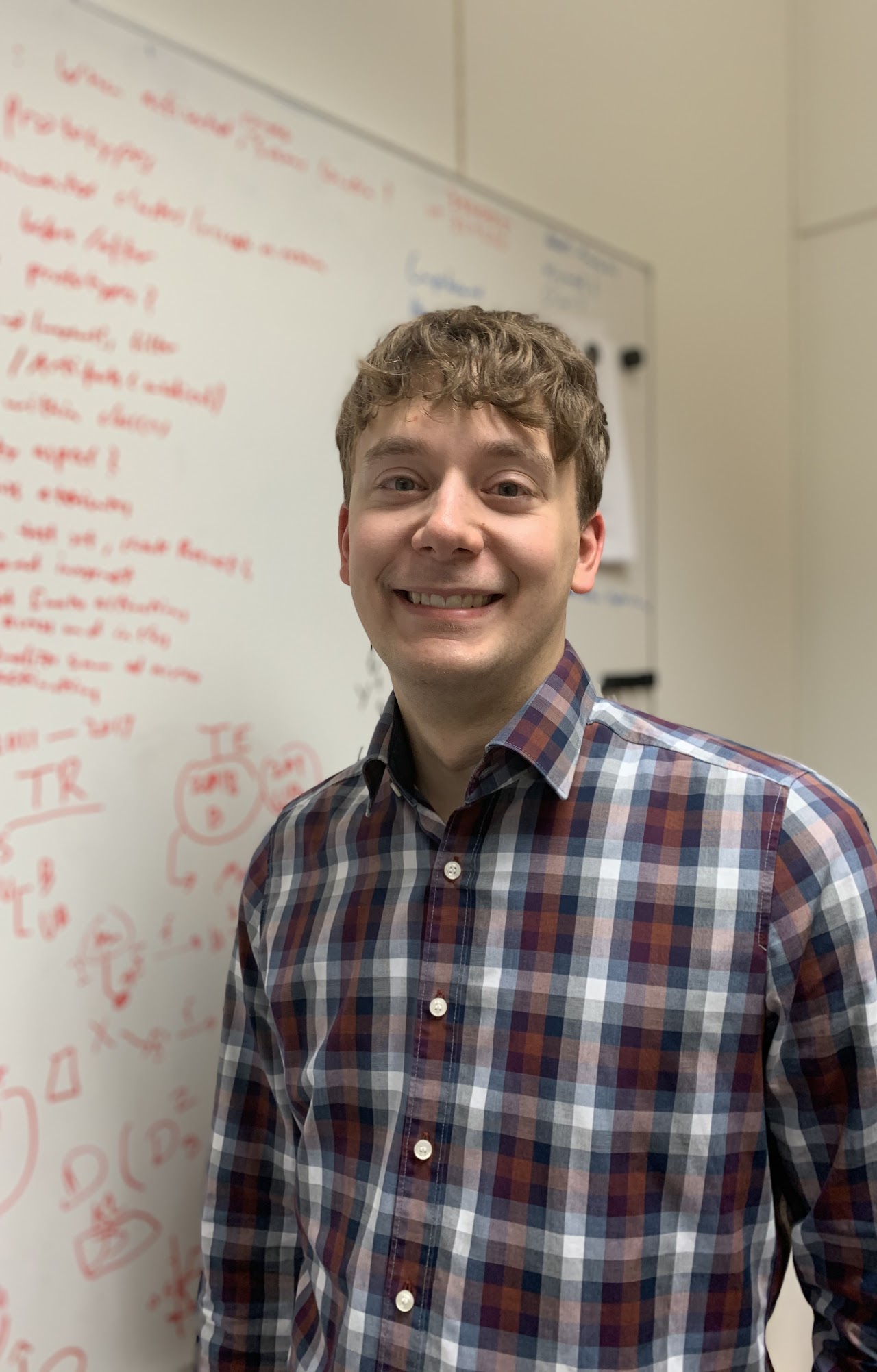}}]{Michael Kampffmeyer} received the Ph.D.\ degree from UiT The Arctic University of Norway, Tromsø, Norway, in 2018. He is an Associate Professor and Head of the Machine Learning Group at UiT The Arctic University of Norway. In addition he is a Senior Researcher at the Norwegian Computing Center. His research interests include the development of deep learning algorithms that learn from limited labeled data and their interpretability.

\end{IEEEbiography}

\begin{IEEEbiography}[{\includegraphics[width=1in,height=1.25in,clip,keepaspectratio]{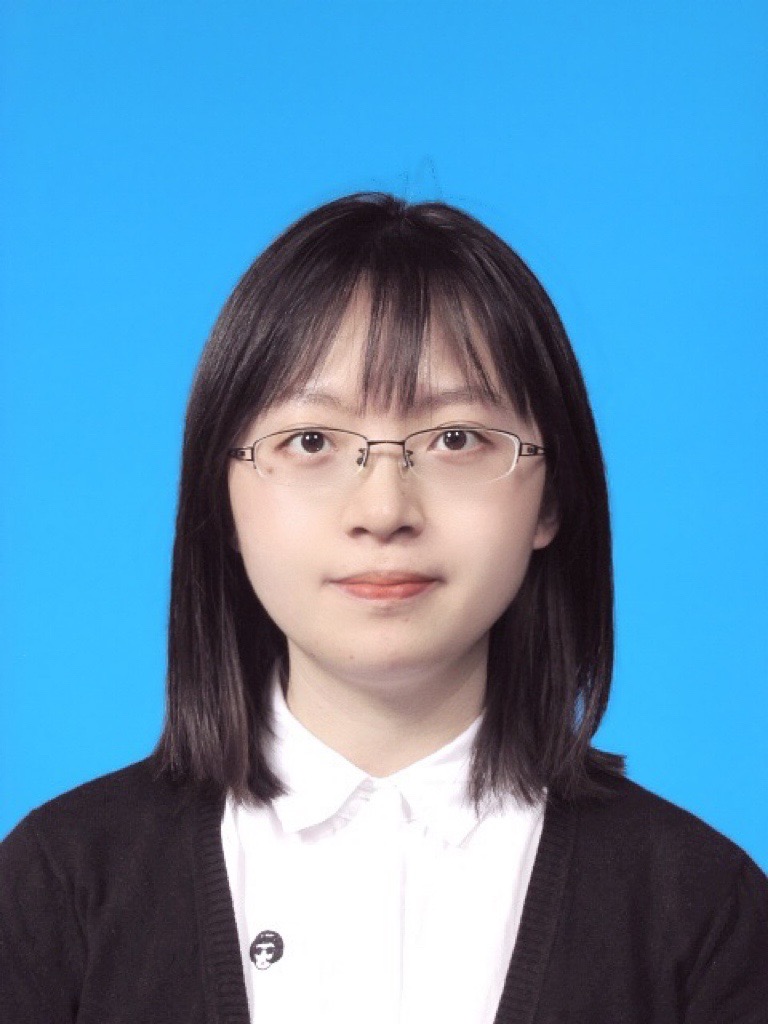}}]{Xin Dong} is currently a researcher at ByteDance. She received the BE degree from Tsinghua University in 2016 and ME degree in 2019. Her research interests include image synthesis, pose estimation and image segmentation.

\end{IEEEbiography}

\begin{IEEEbiography}[{\includegraphics[width=1in,height=1.25in,clip,keepaspectratio]{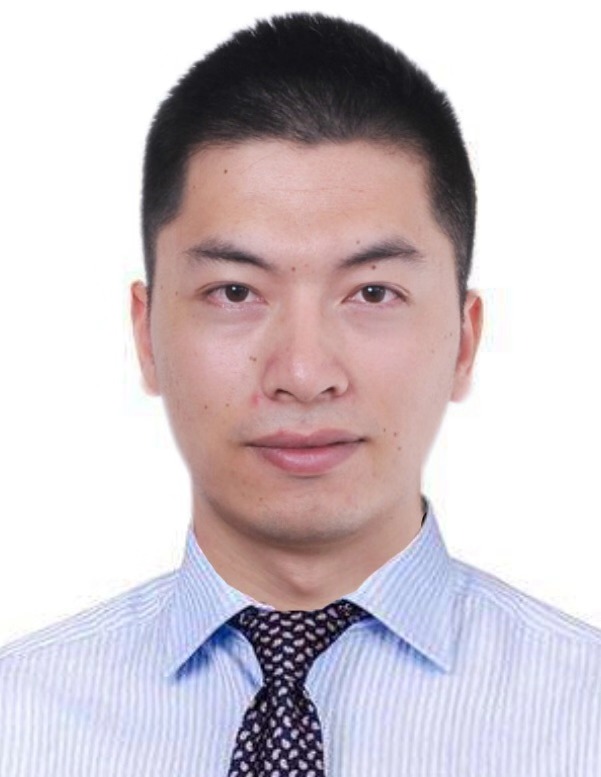}}]{Feida Zhu} is currently a researcher at ByteDance. He received the B.Eng. degree from The University of Science and Technology of China in 2014, and Ph.D degree from The University of Hong Kong in 2018. He worked as a postdoctoral research fellow in the School of Electrical and Electronic Engineering, Nanyang Technological University in 2019. His research interests include image processing, computer vision and machine learning.

\end{IEEEbiography}

\begin{IEEEbiography}[{\includegraphics[width=1in,height=1.25in,clip,keepaspectratio]{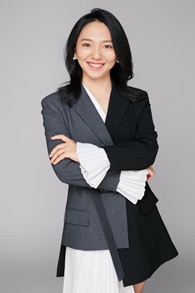}}]{Xiaodan Liang} is currently an Associate Professor at Sun Yat-sen University. She was a postdoc researcher in the machine learning department at Carnegie Mellon University, working with Prof. Eric Xing, from 2016 to 2018. She received her Ph.D. degree from Sun Yat-sen University in 2016. She has published several cutting-edge projects on human-related analysis, including human parsing, pedestrian detection, and instance segmentation, 2D/3D human pose estimation, and activity recognition.

\end{IEEEbiography}







\end{document}